\documentclass[preprint,12pt,3p]{elsarticle}
\usepackage{graphicx}%
\usepackage{multirow}%
\usepackage{amsmath,amssymb,amsfonts}%
\usepackage{amsthm}%
\usepackage{mathrsfs}%
\usepackage[title]{appendix}%
\usepackage{xcolor}%
\usepackage{textcomp}%
\usepackage{manyfoot}%
\usepackage{booktabs}%
\usepackage{algorithm}%
\usepackage{algorithmicx}%
\usepackage{algpseudocode}%
\usepackage{listings}%
\usepackage{longtable}%

\usepackage{amssymb}

\newcommand{\fakepar}[1]{\vspace{3mm}\noindent\textbf{#1}}

\journal{}
\allowdisplaybreaks

\begin{document}

\begin{frontmatter}

\title{Anomaly Detection in Smart Power Grids with Graph-Regularized MS-SVDD: a Multimodal Subspace Learning Approach}

\address[label1]{Faculty of Information Technology and Communication Sciences, Tampere University, FI-33720 Tampere, Finland}
\address[label2]{Technical University of Darmstadt, Karolinenplatz 5 Darmstadt, 64289 Darmstadt, Germany}

\author[label2]{Thomas Debelle}
\cortext[cor1]{corresponding author}
\ead{thomas.debelle@tu-darmstadt.de}
\author[label1]{Fahad Sohrab\corref{cor1}}
\ead{fahad.sohrab@tuni.fi}
\author[label1]{Pekka Abrahamsson}
\ead{pekka.abrahamsson@tuni.fi}

\author[label1]{Moncef Gabbouj}
\ead{moncef.gabbouj@tuni.fi}

\begin{abstract}
Anomaly detection in smart power grids is a critical challenge due to the complexity, heterogeneity, and dynamic nature of sensor data streams. Existing one-class classification methods, particularly Subspace Support Vector Data Description (SVDD), have been extended to multimodal scenarios but often fail to fully exploit the structural dependencies across modalities, limiting their robustness in real-world applications. In this paper, we address this gap by proposing a generalized Multimodal Subspace Support Vector Data Description (MS-SVDD) model with graph-embedded regularization. The method projects data from multiple modalities into a shared low-dimensional subspace while preserving modality-specific structure through Laplacian regularizers. Our approach is evaluated on a three-modality dataset derived from smart grid event time series, using a dedicated preprocessing pipeline for constructing one-class classification training samples. The results demonstrate that our graph-embedded MS-SVDD improves robustness of event detection compared to conventional approaches, highlighting the potential of integrating graph priors with multimodal subspace learning for advancing anomaly detection in critical infrastructure. More broadly, this work contributes to the wider field of AI by illustrating how relational and structural information can be systematically embedded into one-class models, enabling robust learning under complex, high-dimensional, and multimodal conditions.
\end{abstract}


\end{frontmatter}

\section{Introduction}

The use of decarbonized energies in power grids has been widely promoted during the last decade. To reduce fossil fuel dependence and respect the new regulations on carbon emissions, public and private interests have increased investments in renewable energies such as wind, solar, or hydraulic power. Planned energy scenarios consider a significant increase in modern renewable energy supply, whose share could reach $17\%$ by 2030 and $25\%$ by 2050 \cite{Damodaram,choobdari2024robust}.

Despite the expected benefits of these new sources for the environment, their effective deployment in power grids presents new challenges that need to be addressed to ensure stability for end-user distribution. Most renewable energies are, in fact, subject to high variability and seasonality. For instance, in Portugal, wind can increase by $45\%$ in winter and decrease by $45\%$ in summer, leading to a variation in wind power of nearly $100$ kWh \cite{RUSSO2023136997}. Solar power generation is highly influenced by solar radiation and cloud cover, making it less predictable during cloudy days \cite{Berraies2021}. Hydraulic power depends on water availability, which is affected by precipitation, snowmelt, and evaporation rate \cite{Penmetsa2019}. Due to their stochastic nature, these environment-dependent energy sources are less dispatchable than fossil fuel generators, which increases the risk of disturbances in the power grid \cite{Aula2013, zheng2021psml,zuhaib2025identification}. If no action is taken, irregularities caused by supply variations can lead to fluctuations and faults, ultimately resulting in system collapse \cite{rajak2025multiobjective}. It is therefore critical to detect and patch these power anomalies as soon as possible to ensure the continuity of power grid operations, which is vital for most human activities. 

Smart power grids are of great help in monitoring these anomalies. In opposition to traditional one-way power grids, which only carry power from generators to end-users, the smart power grid uses two-way flows of electricity and information to adjust and automate energy production and distribution \cite{6099519}. In Europe and North America, power grids are managed by System Operators (SOs) \cite{POLLITT201232} whose goal is to maintain grid stability by facilitating the market between energy producers and distributors, predicting the energy demand, and managing the grid in real-time \cite{life_cycle_cost_assessment}. A smart power grid allows them to leverage power data through algorithms to increase energy efficiency by adapting production to huge demand fluctuations, forecasting energy demand by identifying seasonal behaviors in measurement profiles, or ensuring grid stability by preventing faults before damage occurs. 

Smart power grids can be seen as three interacting systems: infrastructure, management, and protection systems. An anomaly in a smart power grid is processed as follows: first, data are collected through smart meters, sensors, or Phase Measurement Units, which are parts of the infrastructure system. These data are then carried through the grid to the protection system, where algorithms are run to determine whether an anomaly has occurred. If an anomaly is detected, its type is classified, and its root is localized to provide decision support for SOs ultimately. 

Smart power grids usually provide us with several types of measurements, such as voltage, current, power, frequency, or power factor. All these quantities are correlated, but we cannot determine in advance which one will be the most critical to monitor for anomalies. Thus, we adopt a multimodal learning approach where the goal is to learn from heterogeneous, connected, and interacting data while guaranteeing the best information representation \cite{Liang2023101}. By considering each electrical quantity as a modality, we can leverage multiple types of information and determine which modality is prevalent for anomaly detection.

The detection of these anomalies is a One-Class Classification (OCC) problem, where we aim to determine whether or not a specific instance belongs to a target class. This type of problem can be solved with a machine-learning approach, by constructing a model trained on the positive class instances \cite{YU201937, yang2019one}. The trained model can then be used to distinguish normal measurements from anomalies and is especially suitable to address highly imbalanced datasets, where the number of instances in the positive class is much smaller than in the negative class \cite{sohrab2020ellipsoidal,tsai2021feature}. 

In this paper, we apply Multimodal Subspace Support Vector Data Description (MS-SVDD) to anomaly detection in smart power grids. This model maps multimodal data from high-dimensional feature spaces to a low-dimensional subspace, optimized for OCC \cite{SOHRAB2021107648}. We improve MS-SVDD by proposing and validating new multimodal regularization strategies, which use graph-embedded information to enhance modeling of the shared subspace. We also demonstrate that MS-SVDD can generalize to any number of modalities by incorporating more than two, and use decision strategies to identify which modalities are prevalent in our application. 
We finally evaluate the earliness of the best model configurations to ensure that our model meets specific time constraints. Implementations of the proposed framework are available online in GitHub\footnote{https://github.com/thomas-debelle/mssvdd-smart-grid}.

\section{Background and related works}
In this section, we discuss advancements in anomaly detection and multimodal learning for smart power grids. We first explore early anomaly detection methods, focusing on detecting, classifying, and localizing anomalies to ensure grid stability. Then, we introduce the MS-SVDD, which enhances one-class classification by optimizing a shared subspace for data from multiple modalities.
\subsection{Early anomaly detection in smart power grids}
\begin{figure}[t]
	\centering
    \includegraphics[scale=0.25]{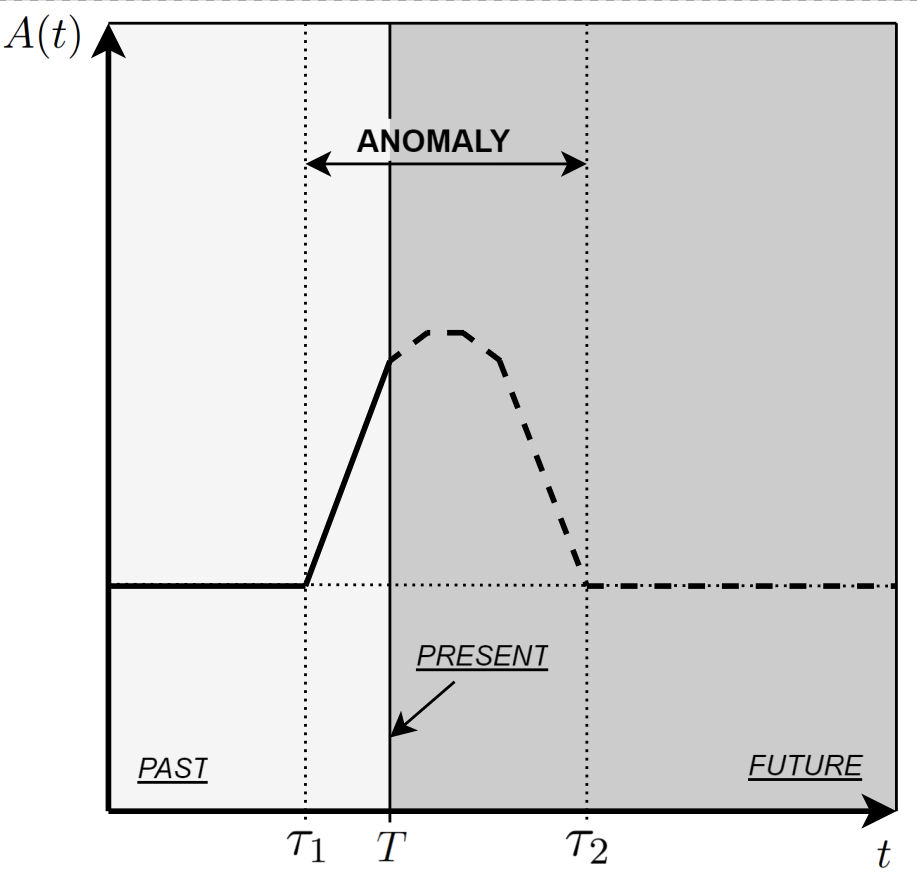}
	\caption{Anomaly detection over an incomplete time series, where $\tau_1$ is the beginning of the anomaly, $\tau_2$ is the end of the anomaly, and $T$ is the current time.}
	\label{fig:early_detection}
\end{figure}

Anomaly detection is only a small part of the whole anomaly management process and should be regarded within the frame of a more general framework. In \cite{zheng2021psml}, the authors considered three critical questions issued by the System Operators to handle events and anomalies: (i) When is an event happening? (ii) What type of event is happening? (iii) Where is the source that caused this event?  
These questions set up a comprehensive procedure by splitting anomaly management into three steps:

\begin{enumerate}
    \item Detecting that an anomaly occurs as soon as possible: Event Detection, or ED.
    \item Classifying the event that occurs as a type of anomaly: Event Classification, or EC.
    \item Localizing the cause of the event in the whole power grid: Event Localization, or EL.
\end{enumerate}
In our work, we only consider the ED problem, which depends on early anomaly detection. Early classification of time series, or early anomaly detection, is a specific OCC problem where the primary goal is to classify an incomplete time series as soon as possible while ensuring a good level of accuracy \cite{9207873}. In other words, if $\tau_1$ is the starting time of the anomaly, $\tau_2$ its ending time, and $T$ its current time, then we aim to classify correctly a time series as abnormal while minimizing $T - \tau_1$ (see Fig. \ref{fig:early_detection}). We should also ensure that the instant of classification $T$ respects the constraint $\tau_1 \leq T \leq \tau_2$. Otherwise, the classification would be considered a false trigger.

The quality of early classification is evaluated through two important metrics:  \textbf{earliness} and \textbf{reliability}. Guaranteeing the best efficiency therefore implies maximizing these two metrics while considering several constraints
\begin{itemize}
    \item Earliness often increases at the cost of reliability: a very sensitive model provides a good earliness but often generates false positives.
    \item Power grid anomalies can have many sources: branch or bus tripping, branch or bus faults, short circuits, and transient failures. This diversity leads to several measurement profiles and makes it harder to find a versatile model to detect all types of anomalies.
    \item To ensure reliability, the model has to be robust to noise. Noisy measurements should not lead to false positives.
    \item The huge number of sensors and quantities measured in the infrastructure system also makes the problem highly dimensional. We thus need to address the curse of dimensionality when solving this OCC problem.
\end{itemize}

Considering the critical need for stability in power grids and the challenges posed by the stochastic nature of new energy sources, early anomaly detection in electrical systems has been an important field of study in the literature. In \cite{6808416}, the authors used a dimensionality reduction approach on synchrophasor data to set up an online application to detect events on a power grid. In \cite{SABER20201113}, a threshold-free searching scheme has been proposed for faulted branch identification in multi-end lines. 

When a disturbance is detected, it has to be cleared in a specific time before the system becomes unstable. This limit is called Critical Clearing Time (CTT) and ranges between $100$ ms and $400$ ms depending on the estimations \cite{zheng2021psml} \cite{Wu2019} \cite{Banjar-Nahor20181183}. Evaluating the earliness of a model enables us to determine if an anomaly can be cleared within the CCT or not. 

\subsection{Multimodal Subspace Support Vector Data Description and applications}

\begin{figure*}[t]
    \centering
    \includegraphics[scale=0.55]{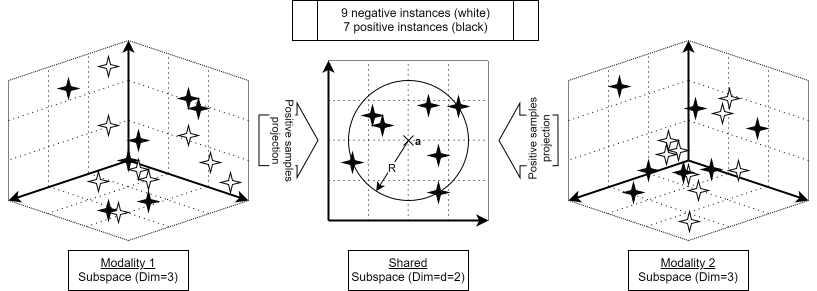}
    \caption{Illustration of MS-SVDD with two modalities. The shared subspace is constructed using the positive instances' information.}
    \label{fig:mssvdd}
\end{figure*}

MS-SVDD is based on Support Vector Data Description (SVDD). This model aims to find a hypersphere of minimal radius that separates target class instances from outliers \cite{Tax2004SupportVD}. Several extensions of SVDD, including fuzzy variants, have improved its robustness for OCC tasks \cite{FSVDD2011}. The process has been first improved through subspace learning by projecting data into a lower-dimensional space where we can find a better boundary, leading to Subspace Support Vector Data Description, or S-SVDD \cite{8545819}. Multimodal learning has then been leveraged to find a shared subspace that optimizes the boundary across different modality spaces (see Fig. \ref{fig:mssvdd}). The MS-SVDD formulation was initially introduced in \cite{SOHRAB2021107648} and later extended with additional regularization strategies, such as distribution entropy regularization for anomaly detection \cite{wang2026distribution} and joint consistency–complementarity regularization for multimodal data \cite{wang2024consistency}.

Let us assume that the elements to be modeled lie in $M$ different modalities of dimensionality $D_m \in \mathbb{N^{*}}$, with $m \in \{1, ..., M\}$. Each instance is represented in all the $M$ modalities, and the set of instances into one modality is described as $\mathbf{X}_m = [\mathbf{x}_{m,1}, \mathbf{x}_{m,2}, ... \mathbf{x}_{m,N}]$, with $\mathbf{x}_{m,i} \in \mathbb{R}^{D_m}$ the instance $i$ in modality $m$, and $N$ the number of instances in the dataset.
The goal of the MS-SVDD algorithm is to find a projection matrix $\mathbf{Q}_m \in \mathbb{R}^{d \times D_m}$ for each modality, in such a way that we can project every instance into a shared $d$-dimensional subspace optimized for OCC. This projection is done using  
\begin{equation}\label{eq:Y_i}
\mathbf{y}_{m,i} = \mathbf{Q}_m \mathbf{x}_{m,i} \:\:,\forall m \in \{1,\dots,M\} \:\:, \forall i \in \{1,\dots,N\},
\end{equation}
where $\mathbf{Y}_m = [\mathbf{y}_{m,1}, \mathbf{y}_{m,2}, ... \mathbf{y}_{m,N}]$ is the matrix containing all the instances of modality $m$ projected into the shared $d$-dimensional subspace.
We aim to minimize the volume of the hypersphere constructed around the training data in the $d$-dimensional subspace, under the constraint that most of the instances must lie within the hypersphere, i.e.
\begin{equation}\label{eq:F_R_a}
\min F(R,\textbf{a}) = R^2 + C\sum_{m=1}^{M}\sum_{i=1}^{N} \xi_{m,i}\nonumber
\end{equation}
s.t.
\begin{eqnarray}\label{eq:F_R_a_st}
\| {\mathbf{Q}_m\mathbf{x}_{m,i}} - \mathbf{a} \|^2_{2} \le R^2 + \xi_{m,i},\\
\xi_{m,i} \ge 0,\nonumber \\
\forall m \in \{1,\dots,M\}, \forall i \in \{1,\dots,N\} ,\nonumber 
\end{eqnarray}
where $R$ is the radius, $\mathbf{a} \in \mathbb{R}^d$ is the center of the hypersphere, $\xi_{m,i}$ are slack variables, and $C$ controls the outliers in the training set \cite{10081907}. This optimization problem can be solved with a gradient descent by updating $\mathbf{Q}_m$ as follows
\begin{equation}\label{eq:Q_m_update}
\mathbf{Q}_m \leftarrow \mathbf{Q}_m - \eta \Delta L_m,
\end{equation}
where $\Delta L_m$ is the gradient of the Lagrangian of Eq. \eqref{eq:F_R_a_st} for modality $m$. The Lagrangian is defined as
\begin{align}
L &= \sum_{m=1}^M \sum_{i=1}^N \alpha_{m,i} \mathbf{x}_{m,i}^T \mathbf{Q}_m^T \mathbf{Q}_m \mathbf{x}_{m,i} \nonumber \\
&\quad - \sum_{m=1}^M \sum_{i=1}^N \sum_{n=1}^M \sum_{j=1}^N \alpha_{m,i} \mathbf{x}_{m,i}^T \mathbf{Q}_m^T \mathbf{Q}_n \mathbf{x}_{n,j} \alpha_{n,j}
+ \beta \omega \label{eq:L}
\end{align}
and we calculate its derivative with 
\begin{align}
\Delta L_m &= \frac{\partial L}{\partial \mathbf{Q}_m} 
= 2 \sum_{i=1}^{N} \alpha_{m,i} \mathbf{Q}_m \mathbf{x}_{m,i} \mathbf{x}_{m,i}^T \nonumber \\
&\quad - 2 \sum_{i=1}^{N} \sum_{j=1}^{N} \sum_{n=1}^{M} \mathbf{Q}_n \mathbf{x}_{n,j} \mathbf{x}_{m,i}^T \alpha_{m,i} \alpha_{n,j}
+ \beta \Delta \omega \label{eq:Delta_L}
\end{align}
$\boldsymbol{\alpha} \in \mathbb{R}^{M \times N}$ is a matrix containing the Lagrangian coefficients, $\Delta \omega$ is the derivative of the regularization term with respect to $\mathbf{Q}_m$, and $\beta$ is a regularization parameter which controls the significance of this term. 

The regularization term enforces smoothness in the learned projection matrices 
$\mathbf{Q}_m$, preventing abrupt variations that could lead to overfitting. In the MS-SVDD setting, this constraint also promotes consistency between modalities while preserving their shared underlying structure. $\omega$ embeds the covariance of data from different modalities in the $d$-dimensional subspace, and is expressed in its general form as
\begin{equation}\label{eq:w_0123}
\omega = \sum_{m=1}^{M}  \text{tr}(\mathbf{Q}_m\mathbf{X}_m \boldsymbol{\nu}_m \boldsymbol{\nu}^T_m \mathbf{X}^T_m\mathbf{Q}^T_m),
\end{equation}
where $\boldsymbol{\nu}_m \in \mathbb{R}^N$ is a vector constructed from elements of $\boldsymbol{\alpha}_m$ with Eq. \eqref{eq:nu_1}, \eqref{eq:nu_2}, \eqref{eq:nu_3}, or \eqref{eq:nu_4}.
An alternative formulation allows us to cross modalities two-by-two such as
\begin{equation}\label{eq:w_456}
\omega_c = \sum_{m=1}^{M} \sum_{n=1}^{M}  \text{tr}(\mathbf{Q}_m\mathbf{X}_m \boldsymbol{\nu}_m \boldsymbol{\nu}^T_n \mathbf{X}^T_n\mathbf{Q}^T_n).
\end{equation}
By setting up different values for $\mathbf{\nu}_m$, we can create several regularizers for the training of the model. To create our different regularizers with $m \in \{1, ..., M\}$, we use
\begin{eqnarray}
\boldsymbol{\nu}_m = \mathbf{0}_N \label{eq:nu_1}\\
\boldsymbol{\nu}_m = \mathbf{1}_N \label{eq:nu_2}\\
\boldsymbol{\nu}_m = \boldsymbol{\alpha}_m \label{eq:nu_3}\\
\boldsymbol{\nu}_m = \boldsymbol{\lambda}_m\label{eq:nu_4}
\end{eqnarray}
where $\mathbf{0}_N$ is a null vector of size $N$, $\mathbf{1}_N$ is a vector of size $N$ filled with $1$ and $\boldsymbol{\lambda}_m$ is a vector having the elements of $\boldsymbol{\alpha}_m$ that are smaller than C, with values corresponding to the outliers ($\alpha_{m,i}>C$) replaced by zeros. We note that with $M=1$, Eq. (\ref{eq:w_0123}) and (\ref{eq:w_456}) are equivalent to the regularizers of S-SVDD.

We implement the Non-linear Projection Trick (NPT) as an alternative to the kernel trick for non-linear data description \cite{6584012}. For each modality, the NPT kernel matrix is calculated as follows
\begin{equation}\label{eq:kernel_npt}
[\mathbf{K}_{m}]_{ij} = \exp  \left( \frac{ -\| \mathbf{x}_{m,i} - \mathbf{x}_{m,j}\|_2^2 }{ 2\sigma^2 } \right),
\end{equation}
where $\sigma$ is a hyperparameter scaling the distance between the instances $\mathbf{x}_{m, i}$ and $\mathbf{x}_{m, j}$. This kernel matrix is then centralized with
\begin{equation}\label{eq:kernel_npt_centr}
    \hat{\mathbf{K}}_m = (\mathbf{I}_N - \frac{1}{N}\mathbf{1}_N\mathbf{1}^T_N)\mathbf{K}_m(\mathbf{I} - \frac{1}{N}\mathbf{1}_N\mathbf{1}^T_N),
\end{equation}
where $\mathbf{1}_N \in \mathbb{R}^N$ is a vector full of ones and $\mathbf{I}_N$ is the identity matrix of size $N\times N$. We can finally compute the matrix containing the data in the kernel space as
\begin{equation}\label{eq:kernel_space}
    \boldsymbol{\Phi}_m = (\mathbf{A}_m^{\frac{1}{2}})^{+}\mathbf{U}_m^{+}\mathbf{U}_m\mathbf{A}_m\mathbf{U}_m^T,
\end{equation}
where we consider the eigendecomposition $\hat{\mathbf{K}}_m = \mathbf{U}_m \mathbf{A}_m \mathbf{U}_m^T$. Here, $\mathbf{A}_m$ contains the non-negative eigenvalues of the centered kernel matrix, $\mathbf{U}_m$ contains the corresponding eigenvectors, and the $+$ exponent denotes the Moore-Penrose pseudo-inverse. We can then use the matrix $\boldsymbol{\Phi}_m$ as training data instead of $\mathbf{X}_m$ during the training phase. In the same way, for the testing phase, a centralized kernel vector $\boldsymbol{\Phi}_{m*}$ is computed for each instance $\mathbf{x}_{m*} \in \mathbb{R}^{D_m}$ with
\begin{align}
\boldsymbol{\Phi}_{m*} &= \left( \boldsymbol{\Phi}_m^T \right)^{+} 
\left( \mathbf{I}_N - \frac{1}{N} \mathbf{1}_N \mathbf{1}_N^T \right) \nonumber \\
&\quad \times \left( \mathbf{K}_{m*} - \frac{1}{N} \mathbf{K}_m \mathbf{1}_N \right)
\label{test_npt}
\end{align}

where $\mathbf{K}_{m*}$ is the testing kernel matrix calculated with the test instances using Eq. \ref{eq:kernel_npt}. $\boldsymbol{\Phi}_{m*}$ is finally used as testing data instead of the testing dataset $\mathbf{X}_{m*}$.  

Numerous challenges have been addressed using SVDD and subspace learning \cite{laakom2023convolutional, kilickaya2023hyperspectral, al2024malware}. In \cite{trustworthiness2024}, the trustworthiness of $\mathbb{X}$ users has been evaluated with an S-SVDD approach over social network features, such as the number of friends, the number of followers, or the number of retweets. In \cite{10372038}, authors addressed the problem of credit card fraud detection by leveraging various features over highly imbalanced datasets. Regarding multimodal learning, MS-SVDD has been applied over multiple OCC problems involving two modalities, such as Robot Execution Failures with torque and force, Handwritten characters with Zernike moment and morphological features, or SPECTF dataset with stress and rest condition images \cite{SOHRAB2021107648}. MS-SVDD has also been applied to medical data for early detection of myocardial infarction \cite{10081907,zahid2024refining}. 

The diversity of these different works emphasizes the versatility of S-SVDD and MS-SVDD in terms of possible applications. Therefore, some directions have not been explored yet. For instance, multimodal applications were always limited in practice to two modalities. Also, as mentioned earlier, several improvements, such as graph embedding, were only developed for unimodal S-SVDD and not MS-SVDD. In this article, we aim to extend all these improvements to MS-SVDD, and to apply this model to more than two modalities.

\section{Graph regularized multimodal subspace support vector data description}
\subsection{Multimodal graph regularizers}

Besides proposing a novel application of MS-SVDD for early detection, we generalize graph-embedded regularizers to any number of modalities. Graph-embedded regularizers allow us to leverage distance information in the original feature spaces to improve the regularization process. In a multimodal context, graph embedding is performed separately for each modality and then aggregated in the graph-regularizer calculation.

\begin{figure}[t]
    \centering
    \includegraphics[width=1.0\linewidth]{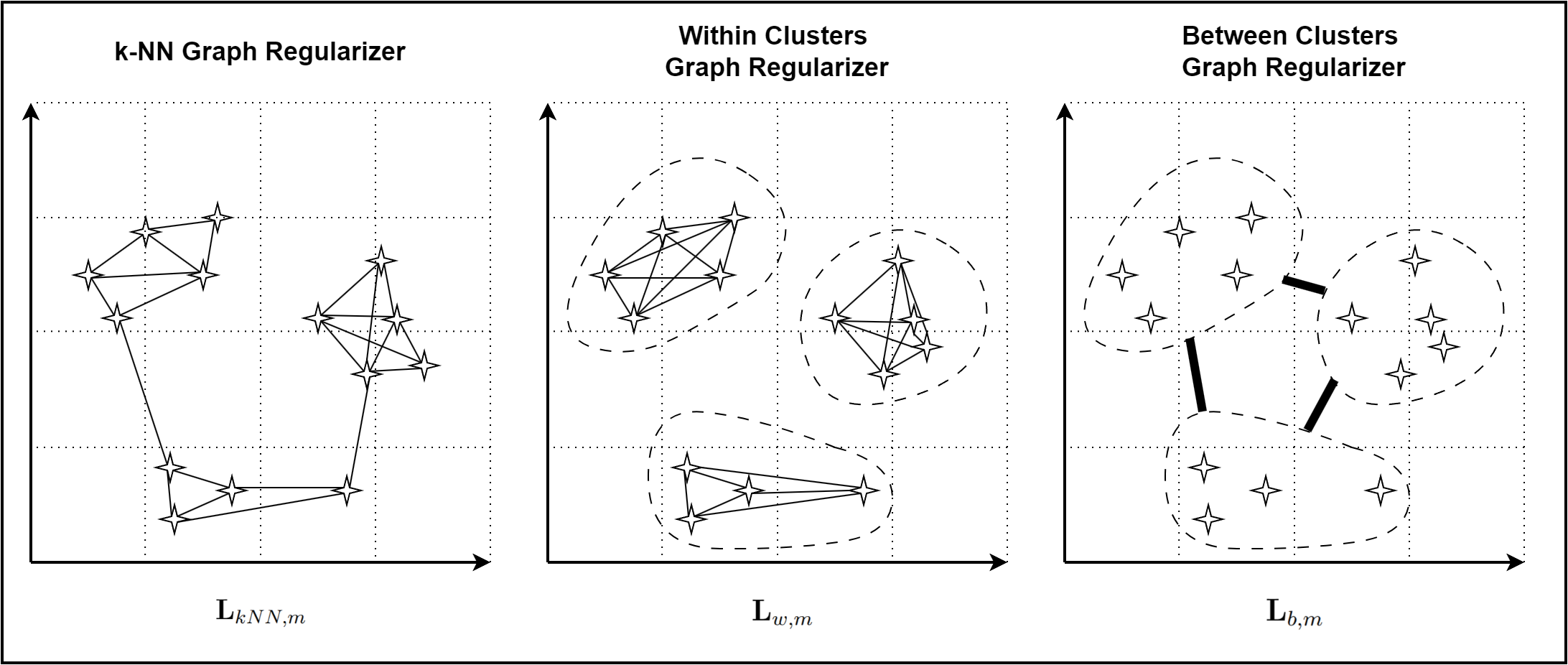}
    \caption{High-level illustration of graph-embedded regularizers for a modality $m$ with $D_m=2$ and $k=3$. Each plot corresponds to one type of Laplacian matrix.}
    \label{fig:graph_embedded_reg}
\end{figure}

Let us give a new expression for the regularization term that leverages graph information from each modality in the lower-dimensionality space
\begin{equation}\label{eq:ge_reg}
\omega_g = \sum_{m=1}^{M}  \text{tr}(\mathbf{Q}_m\mathbf{X}_m \boldsymbol{L}_{g,m} \mathbf{X}^T_m\mathbf{Q}^T_m),
\end{equation}
where $\mathbf{L}_{g,m} \in \mathbb{R}^{N\times N}$ is the Laplacian matrix of the graph $g$.
Given a specific type of Laplacian matrix, a graph is constructed for every modality based on the distance information between all the instances. We propose three different graph Laplacians: local geometric information Laplacian with k-Nearest-Neighbors (k-NN) $\mathbf{L}_{kNN,m}$, within-cluster information Laplacian $\mathbf{L}_{w,m}$, and between-cluster information Laplacian $\mathbf{L}_{b,m}$. A high-level illustration of these graphs is shown in Fig. \ref{fig:graph_embedded_reg}.

The k-NN information Laplacian is defined as
\begin{equation}\label{eq:L_knn}
\mathbf{L}_{kNN,m} = \mathbf{D}_{kNN,m} - \mathbf{A}_{kNN,m}
\end{equation}
where $[\mathbf{A}_{kNN,m}]_{ij} = 1$ if $\mathbf{x}_i \in \mathcal{N}_{m,i} \cup \mathcal{N}_{m,j}$, $0$ otherwise, with $\mathcal{N}_{m,i}$ denoting the nearest neighbors of $\boldsymbol{x}_i$ for modality $m$. For each instance in every modality $m$, connections are constructed with the $k$ nearest neighbors. These connections represent the relative proximity of the instances in the feature space.

The within-clusters information Laplacian is defined as
\begin{equation}\label{eq:L_w}
\mathbf{L}_{w,m} = \mathbf{I}-\sum_{c \in \mathcal{C}_m} \frac{1}{N_c} \mathbf{1}_c \mathbf{1}^T_c,
\end{equation}
where $\mathbf{I}$ is an identity matrix, $\mathcal{C}_m$ is the set of clusters for modality $m$, $N_c$ is the number of instances belonging to the cluster $c$, and $\mathbf{1}_c$ is a vector where the entries are ones for instances belonging to cluster $c$ and zeros otherwise. With this type of graph, $k$ clusters are formed for every modality $m$, and the instances corresponding to the same cluster are all connected. 

The between-clusters information Laplacian is finally defined as
\begin{equation}\label{eq:L_b}
\mathbf{L}_{b,m} = \sum_{c \in \mathcal{C}_m} N_c (\frac{1}{N_c} \mathbf{1}_c - \frac{1}{N} \mathbf{1})(\frac{1}{N_c} \mathbf{1}_c - \frac{1}{N} \mathbf{1}^T)
\end{equation}
where $N$ is the total number of instances and $\mathbf{1}$ is a vector full of ones. Within these graphs, $k$ clusters are formed for every modality $m$, and the instances that are not in the same clusters are connected.  

By minimizing these three graph-embedded regularizers in the optimization process, we are now able to leverage distance information for regularization. The full graph-regularized MS-SVDD process is available in Algo. \ref{algo:train} and Algo. \ref{algo:test}.

\subsection{Decision strategies} \label{sec:decision_strategies}
At the end of the testing process, class predictions are computed for each modality and then fused to determine the final class assigned to an instance. In our framework, we use decision strategies expressed as Boolean equations, where each modality's classification serves as an input and the final instance class is the output. Let $p_m \in \mathbb{B}$ denote the value predicted by MS-SVDD for modality $m$, and let $p$ denote the final value assigned to the instance, with $p_m = 0$ for the non-target class and $p_m = 1$ for the target class. We can then define the following three decision strategies:
\begin{enumerate}
    \item \textbf{AND} strategy, $p=p_1.p_2.p_3$: the instance is assigned the target label if the representations coming from every modality are classified as in the target class. 
    \item \textbf{OR} strategy, $p=p_1+p_2+p_3$: the instance is assigned the target label if at least one of the representations coming from every modality is classified as in the target class.
    \item \textbf{UNI}$_{m}$ strategy, $p=p_m$: the instance is assigned the target label if modality $m$ is classified as in the target class.
\end{enumerate}
Testing different decision strategies on a specific problem allows us to study the modality imbalance, where certain modalities' contributions are suppressed by dominant ones \cite{Han20244591}. We can thus identify which modalities are prevalent without prior information.

\begin{algorithm}[H]
\caption{Training of Graph-Regularized MS-SVDD}
\label{algo:train}
\begin{algorithmic}[1]

\Require $\mathbf{X}_m$ for $m=1,\dots,M$; regularizer type; $\beta$, $\eta$, $d$, $C$, $\sigma$, $k$
\Ensure Projection matrices $\mathbf{Q}_m$, radius $R$, Lagrange multipliers $\boldsymbol{\alpha}$

\Statex \textbf{(Optional NPT mapping)}
\If{NPT enabled}
    \For{$m=1$ to $M$}
        \State Compute kernel matrix $\mathbf{K}_m$ using Eq.~\eqref{eq:kernel_npt}
        \State Center it to obtain $\hat{\mathbf{K}}_m$ using Eq.~\eqref{eq:kernel_npt_centr}
        \State Compute $\boldsymbol{\Phi}_m$ using Eq.~\eqref{eq:kernel_space}
        \State Set $\mathbf{X}_m \gets \boldsymbol{\Phi}_m$
    \EndFor
\EndIf

\Statex \textbf{Initialize subspace}
\For{$m=1$ to $M$}
    \State Initialize $\mathbf{Q}_m$ via linear PCA on $\mathbf{X}_m$
\EndFor

\Statex \textbf{Main optimization loop}
\For{$iter = 1$ to $\text{max\_iter}$}
    \For{$m=1$ to $M$}
        \State Compute projected data $\mathbf{Y}_m = \mathbf{Q}_m \mathbf{X}_m$
    \EndFor

    \State Concatenate all $\mathbf{Y}_m$ into shared matrix $\mathbf{Y}$

    \State Solve SVDD in shared subspace → obtain $\boldsymbol{\alpha}, R$ (via dual optimization)

    \For{$m=1$ to $M$}

        \State \textbf{Compute graph Laplacian for modality $m$}
        \If{regularizer = $k$NN}
            \State Construct $\mathbf{L}_{kNN,m}$ using Eq.~\eqref{eq:L_knn}
        \ElsIf{regularizer = within-cluster}
            \State Construct $\mathbf{L}_{w,m}$ using Eq.~\eqref{eq:L_w}
        \ElsIf{regularizer = between-cluster}
            \State Construct $\mathbf{L}_{b,m}$ using Eq.~\eqref{eq:L_b}
        \EndIf

        \State Compute gradient component $\Delta \omega_g$ using Eq.~\eqref{eq:ge_reg}

        \State Compute $\Delta L_m$ using Eq.~\eqref{eq:Delta_L} plus $\beta\,\Delta\omega_g$

        \State Update projection:
        \[
            \mathbf{Q}_m \gets \mathbf{Q}_m + \eta\, \Delta L_m
        \]

        \State Orthogonalize $\mathbf{Q}_m$ with QR decomposition

    \EndFor
\EndFor

\end{algorithmic}
\end{algorithm}

\begin{algorithm}[H]
\caption{Testing of Graph-Regularized MS-SVDD}
\label{algo:test}
\begin{algorithmic}[1]

\Require Test data $\mathbf{X}_{m*}$ for $m=1,\dots,M$; trained $\mathbf{Q}_m$, $R$, $\boldsymbol{\alpha}$
\Ensure Predicted labels $\mathbf{p}_*$

\Statex \textbf{(Optional NPT mapping)}
\If{NPT enabled}
    \For{$m=1$ to $M$}
        \State Compute kernel vector $\boldsymbol{\Phi}_{m*}$ using Eq.~\eqref{test_npt}
        \State Set $\mathbf{X}_{m*} \gets \boldsymbol{\Phi}_{m*}$
    \EndFor
\EndIf

\Statex \textbf{Projection and scoring}
\For{$m=1$ to $M$}
    \State Compute $\mathbf{Y}_{m*} = \mathbf{Q}_m \mathbf{X}_{m*}$
    \State Compute SVDD distance to center $\mathbf{a}$:
    \[
    d_{m*} = \|\,\mathbf{Y}_{m*} - \mathbf{a}\,\|_2^2
    \]
    \State Assign preliminary label:
    \[
        p_{m*} = \begin{cases}
        1 & d_{m*} \le R^2 \\
        0 & \text{otherwise}
        \end{cases}
    \]
\EndFor

\Statex \textbf{Decision fusion}
\State Fuse $\{p_{1*},...,p_{M*}\}$ using the selected strategy (AND / OR / unimodal)
\State Output final label $p_*$

\end{algorithmic}
\end{algorithm}

\subsection{Complexity analysis}\label{sec:complexity}
The proposed method in Algo. \ref{algo:train} has the following main steps:
\begin{enumerate}
    \item Initializing the projection matrix via PCA: by computing the covariance matrix, then the eigenvalue decomposition. The complexity of these steps for all modalities is $\mathcal{O}(\min(N^2 \sum_{\mathcal{D}}, \sum_{\mathcal{D}^2} N) + \sum_{\mathcal{D}^3})$ with $\sum_{\mathcal{D}^n}=\sum_{m=1}^{M} D_m^n$.
    \item Mapping data from all modalities to a shared $d$-dimensional subspace: complexity of multiplying $d \times D_m$ and $D_m \times N$ i.e. $\mathcal{O}(dD_m N)$. Repeating this for every modality, we get $\mathcal{O}(d\sum_{\mathcal{D}} N)$.
    \item Solving SVDD in the shared subspace to obtain $\alpha$: complexity of $\mathcal{O}(M^3 N^3)$ for $N$ data points coming from $M$ different modalities \cite{zheng2016smoothly}. 
    \item Calculating the Laplacian matrix for the selected graph-embedded relations involves a complexity that depends on the chosen Laplacian. We assume that computing the Laplacian has $\leq$ $\mathcal{O}(N^3)$ complexity. 
    \item Updating $\Delta L_m$ for each modality: complexity of $\mathcal{O}(2dN^2(\sum_{\mathcal{D}})^2)$
    \item Updating $Q_m$ for each modality: complexity of $\mathcal{O}(d\sum_{\mathcal{D}})$
    \item Orthogonalizing and normalizing $Q_m$ with QR decomposition: complexity of $\mathcal{O}(d(\sum_{\mathcal{D}^2}))$ for all modalities. 
\end{enumerate}
The full complexity of one training iteration is $\mathcal{O}(\min(N^2\sum_{\mathcal{D}},\sum_{\mathcal{D}^2}N)+\sum_{\mathcal{D}^3}+M^3 N^3)$, which can be reduced to $\mathcal{O}(N^3)$ under certain conditions \cite{SOHRAB2021107648}. As the testing process in Algorithm \ref{algo:test} does not change from the original MS-SVDD, its complexity remains $\mathcal{O}(d \sum_{\mathcal{D}} + Md)$.

\section{Experiments}

\subsection{Dataset and preprocessing}

We conduct our experiments using the synthetic PSML dataset \cite{zheng2021psml}. This dataset has been generated based on transmission and distribution simulations (T+D) and addresses the lack of publicly available data for smart power grids. It includes 550 events in the transmission grid, each composed of 91 4-second Phase Measurement Unit (PMU) measurements: 24 normalized voltage, 32 active power measurements, and 32 reactive power measurements. Although additional variables (e.g., frequency and current) are available in the minute-level data, they do not include labeled disturbances. Therefore, we restrict our study to the millisecond-level measurements.

\begin{figure}[h]
	\centering
	\includegraphics[scale=0.095]{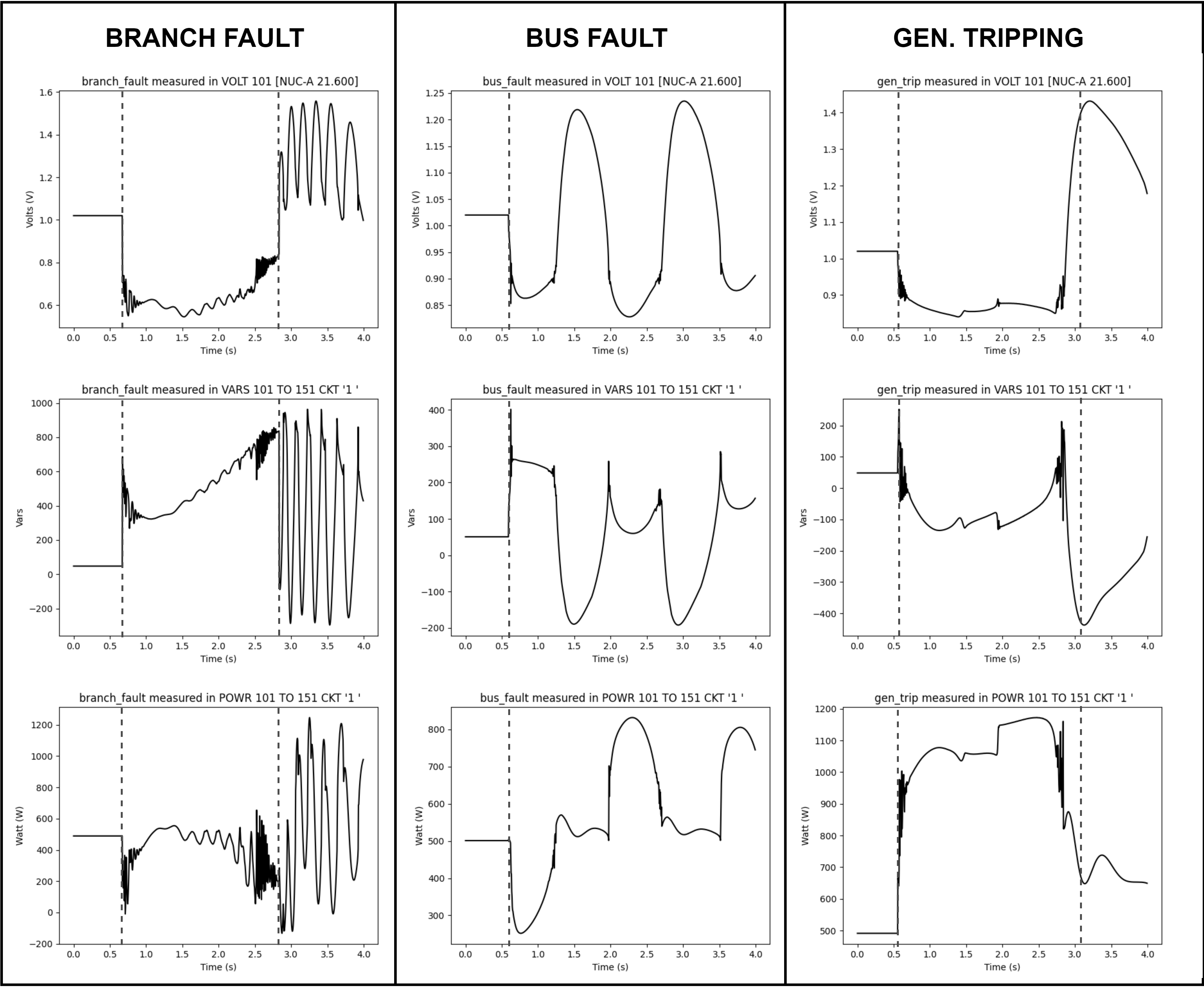}
	\caption{Examples of several time series measured on the same bus while an event is occurring somewhere in the grid. Each row corresponds to an electrical quantity (voltage, reactive power, active power), and the dotted lines are the selected event's beginnings and ends. Note that the signal still oscillates after the end of an event.}        
	\label{fig:events}
\end{figure}

Voltage measurements are normalized and expressed in “per unit" (p.u) with a value near 1 in normal conditions \cite{KIM2019353}, while reactive and active power are respectively expressed in Vars and Watts. Voltage is measured on every bus of the transmission grid while reactive and active powers are measured on the branches connecting these buses. Every event is annotated with an event start time, an event end time, a location in the grid, and an event class: branch fault, branch tripping, bus fault, bus tripping, and generator tripping. Examples of event measurements are shown in Fig. \ref{fig:events}.

For the ED problem, we pursue two objectives: maximizing the reliability of early detection and maximizing the earliness of event triggers. We first focus on identifying the best multimodal OCC model configurations to maximize reliability, and then we evaluate the earliness of these configurations. Ultimately, this allows us to select an appropriate trade-off between the two objectives depending on specific requirements.

As the PSML dataset was not initially made for Event Detection with Multimodal OCC models, we derive it to create our own ED Multimodal Dataset. In this new dataset, each sample is a set of every available measurements at a specific instant. A sample is labeled as abnormal (positive) if its timestamp \(T\) falls within the anomaly interval \([\tau_1, \tau_2]\), and as normal (negative) otherwise. Specifically, a sample is normal if \(T < \tau_1\) or \(T > \tau_2\). Instances with \(T > \tau_2\) are considered normal according to the dataset annotation protocol, which assumes the fault has been cleared and the system has returned to stable conditions. While in practice a fault exceeding the Critical Clearing Time (CCT) may prevent recovery, such cases are not treated as extensions of the same event in the dataset.

\begin{figure*}[t]
	\centering
    \includegraphics[scale=0.28]{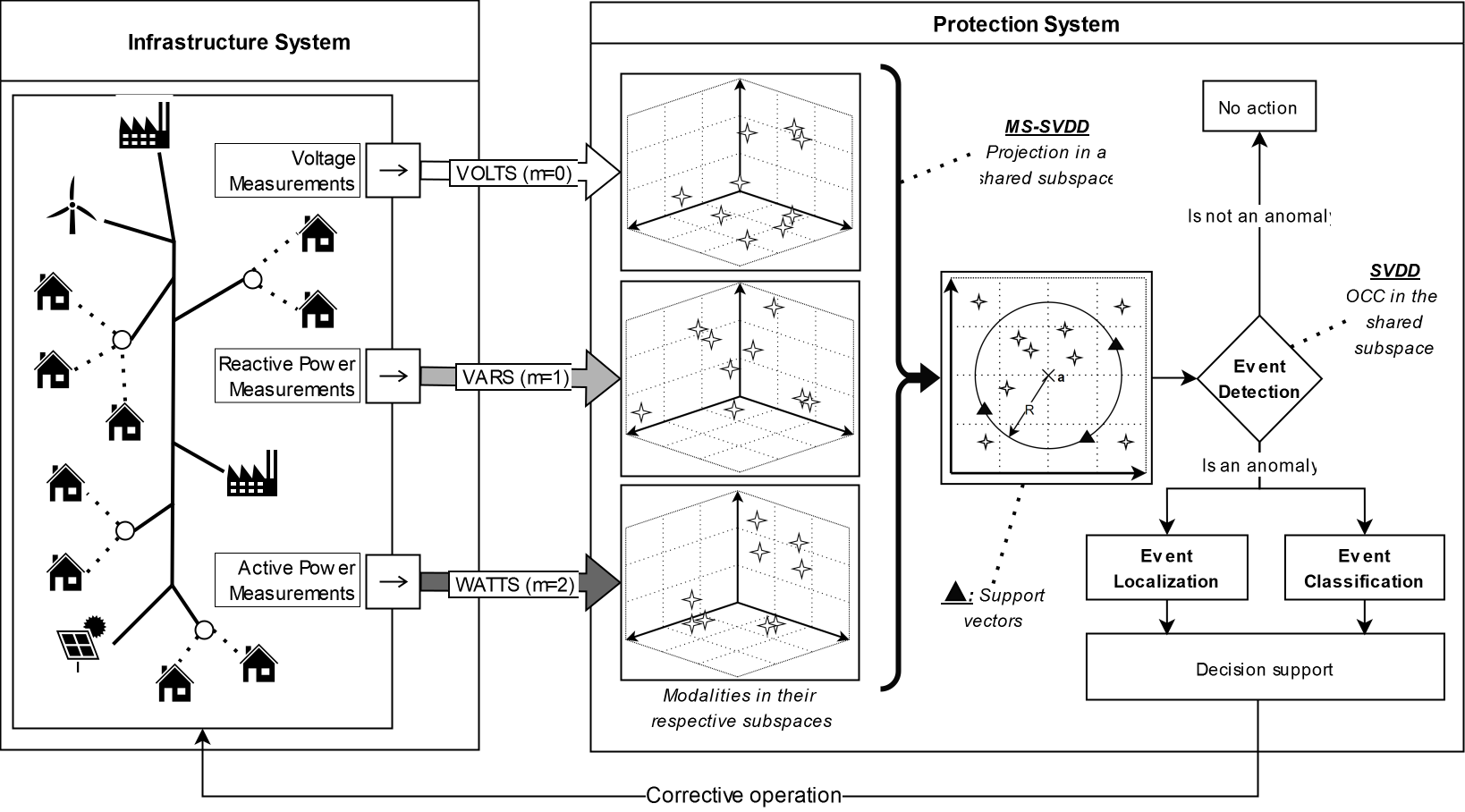}
	\caption{MS-SVDD for Event Detection in Smart Power Grid. Each electrical quantity is considered as one modality lying in its subspace. These data are then projected into a shared subspace, where Event Detection is applied.}
	\label{fig:mssvdd_power_grid}
\end{figure*}

Sample measurements are split between three modalities: voltage (modality 0), reactive power (modality 1), and active power (modality 2). These modality representations are all used by the MS-SVDD algorithm to find the shared subspace for event detection. The original dimensionality of every instance is $D=(23, 34, 34)$ where $D_m$ is the dimensionality of modality $m$. The preprocessing is summarized in Fig. \ref{fig:mssvdd_power_grid}.

For each event in the PSML dataset, we extract 1 positive sample and 1 negative sample. We then randomly dispatch these samples between a training split and a testing split, with $70\%$ of samples in the training split. 
To evaluate the robustness of our model, we add Gaussian noise to each instance by first computing the standard deviation of each PMU measurement within the dataset windows, multiplying it by a predetermined noise factor, and using the resulting value as the standard deviation of a zero-mean normal distribution. Upon loading, the data are finally standardized separately for each modality such as
\begin{eqnarray}\label{eq:data_norm}
    \mathbf{X}_{m*} \leftarrow \frac{\mathbf{X}_{m*} - \mu(\mathbf{X}_{m})}{\sigma(\mathbf{X}_{m})}
\end{eqnarray}
$\forall m \in \{1...M\}$, where $\mathbf{X}_{m}$ is the training split, $\mathbf{X}_{m*}$ is the testing split, $\mu$ calculates a matrix's mean and $\sigma$ calculates a matrix's standard deviation. We present a detailed flow of the different preprocessing steps in the Supplementary Material.

We finally evaluate earliness on selected model configurations using 50 instances from the testing split, with a Critical Clearing Time of 100 ms ($\frac{1}{40}$ of a measurement length). The earliness dataset contains multiple consecutive instances, generated by rolling over an anomaly time series. These instances are then evaluated chronologically by the model: the first instance classified as an anomaly determines the detection delay from the start of the event. Applying this process to every testing event allows us to estimate the average model delay and earliness. In summary, reliability evaluation measures the model's ability to correctly classify randomly selected samples, whereas earliness evaluation assesses its capability to detect real-time anomalies as fast as possible.

\subsection{Experimental setup}
\fakepar{MS-SVDD configurations:} we compare 10 different MS-SVDD regularizers: 4 from the general definition in Eq. \ref{eq:w_0123} ($\omega_0$, $\omega_1$, $\omega_2$, $\omega_3$), 3 from the alternative crossed-modalities definition in Eq. \ref{eq:w_456} ($\omega_4$, $\omega_5$, $\omega_6$) and 3 from the graph-embedded definition in Eq. \ref{eq:ge_reg} ($\omega_7$, $\omega_8$, $\omega_9$). These regularizers are all listed in Table \ref{tab:regs}. In our experiments, these regularizers are evaluated in combination with the other parameters across all the values listed in Table \ref{tab:variants}. Fine-tuning configurations are evaluated using the \textit{Geometric Mean} (Eq. \ref{eq:metrics_gmean}), and we select the MS-SVDD hyperparameters from the following grid of values:
\begin{itemize}
  \item$\beta\in\{10^{-4},10^{-3},10^{-2},10^{-1},10^{0},10^{1},10^{2},10^{3},10^{4}\},$
  \item$C\in\{0.01,0.05,0.1,0.2,0.3,0.4,0.5,0.6\}$,
  \item$d\in\{1,2,3,4,5,10,20\}$,
  \item$\eta=0.1$,
  \item$\sigma\in\{10^{-3},10^{-2},10^{-1},10^{0},10^{1},10^{2},10^{3}\}$ (only with NPT enabled),
  \item$k\in\{1,2,3,4,5,6,7,8,9,10\}$ (only for graph-embedded regularizers).
\end{itemize}

\def\arraystretch{1.5}
\begin{table}[t]
\centering
\small
\begin{tabular}{|l|}
\hline
$\omega_0 = 0$ \\ \hline
$\omega_1 = \sum_{m=1}^{M} \text{tr}(\mathbf{Q}_m\mathbf{X}_m  \mathbf{X}^T_m\mathbf{Q}^T_m)$ \\ \hline
$\omega_2 = \sum_{m=1}^{M}  \text{tr}(\mathbf{Q}_m\mathbf{X}_m \boldsymbol{\alpha}_m \boldsymbol{\alpha}^T_m \mathbf{X}^T_m\mathbf{Q}^T_m)$ \\ \hline
$\omega_3 = \sum_{m=1}^{M} \text{tr}(\mathbf{Q}_m\mathbf{X}_m \boldsymbol{\lambda}_m \boldsymbol{\lambda}^T_m \mathbf{X}^T_m\mathbf{Q}^T_m)$ \\ \hline
$\omega_4 = \sum_{m=1}^{M} \sum_{n=1}^{M} \text{tr}(\mathbf{Q}_m\mathbf{X}_m\mathbf{X}^T_n\mathbf{Q}^T_n)$ \\ \hline
$\omega_5 =\sum_{m=1}^{M} \sum_{n=1}^{M}  \text{tr}(\mathbf{Q}_m\mathbf{X}_m \boldsymbol{\alpha}_m \boldsymbol{\alpha}^T_n \mathbf{X}^T_n\mathbf{Q}^T_n)$ \\ \hline
$\omega_6 =\sum_{m=1}^{M} \sum_{n=1}^{M}  \text{tr}(\mathbf{Q}_m\mathbf{X}_m \boldsymbol{\lambda}_m \boldsymbol{\lambda}^T_n \mathbf{X}^T_n\mathbf{Q}^T_n)$ \\ \hline
$\omega_7 =\sum_{m=1}^{M}  \text{tr}(\mathbf{Q}_m\mathbf{X}_m \boldsymbol{L}_{kNN,m} \mathbf{X}^T_m\mathbf{Q}^T_m)$ \\ \hline
$\omega_8 =\sum_{m=1}^{M}  \text{tr}(\mathbf{Q}_m\mathbf{X}_m \boldsymbol{L}_{w,m} \mathbf{X}^T_m\mathbf{Q}^T_m)$ \\ \hline
$\omega_9 =\sum_{m=1}^{M}  \text{tr}(\mathbf{Q}_m\mathbf{X}_m \boldsymbol{L}_{b,m} \mathbf{X}^T_m\mathbf{Q}^T_m)$ \\ \hline
\end{tabular}
\caption{List of regularizers used for our experiments. $\omega_7$, $\omega_8$ and $\omega_9$ are graph-embedded regularizers.}
\label{tab:regs}
\end{table}
\def\arraystretch{1}

\begin{table}[]
\centering
\small
\begin{tabular}{|l|}
\hline
\textbf{Regularizer type} \\ \hline
\begin{tabular}[c]{@{}l@{}}We evaluate 10 different regularizers. Regularizers 7, 8, 9 \\ correspond to the graph-embedded regularizers, and are \\ compared to the other ones.\end{tabular} \\
\begin{tabular}[c]{@{}l@{}}$\rightarrow$ $\omega_0$, $\omega_1$, $\omega_2$, $\omega_3$, $\omega_4$, $\omega_5$, \\ $\omega_6$, $\omega_7$, $\omega_8$, $\omega_9$\end{tabular} \\ \hline
\textbf{Decision strategy} \\ \hline
\begin{tabular}[c]{@{}l@{}}We use 5 simple decision strategies to study the modality \\ imbalance in our problem.\end{tabular} \\
$\rightarrow$ AND, OR, $\text{UNI}_0$, $\text{UNI}_1$, $\text{UNI}_2$ \\ \hline
\textbf{NPT} \\ \hline
\begin{tabular}[c]{@{}l@{}}We apply the Non-linear Projection Trick to enable non-linear\\ data learning.\end{tabular} \\
$\rightarrow$ 1 (yes), 0 (no) \\ \hline
\textbf{Noise} \\ \hline
\begin{tabular}[c]{@{}l@{}}We apply the model on a noised version of the dataset to evaluate \\ its robustness.\end{tabular} \\
$\rightarrow$ $0\%$, $10\%$ \\ \hline
\end{tabular}
\caption{List of experiment variants describing the possible values for every parameters.}
\label{tab:variants}
\end{table}

\fakepar{Reliability evaluation:} we evaluate reliability using five metrics. The \textit{True Positive Rate} $tpr$ is defined as
\begin{equation}\label{eq:metrics_tpr}
    tpr = \frac{n_{tp}}{N_{p*}},
\end{equation}
where $n_{tp}$ is the number of positive instances classified correctly (true positives) and $N_{p*}$ is the total number of positive instances in the testing split. This rate is also called Sensitivity or Recall, as it measures the reaction of the model to small variations. The \textit{True Negative Rate} $tnr$ is defined as
\begin{equation}\label{eq:metrics_tnr}
    tnr = \frac{n_{tn}}{N_{n*}},
\end{equation}
where $n_{tn}$ is the number of negative instances classified correctly (true negatives) and $N_{n*}$ is the total number of negative instances in the testing split. This rate is also called Specificity, as it measures the inertia of the model when facing small variations. From $tpr$ and $tnr$, we define the \textit{Accuracy} $acc$ as
\begin{equation}\label{eq:metrics_accu}
    acc= \frac{n_{tp} + n_{tn}}{N_{p*} + N_{n*}},
\end{equation}
and the \textit{Precision} $pre$ as
\begin{equation}\label{eq:metrics_pre}
    pre=\frac{n_{tp}}{n_{tp}+n_{fp}},
\end{equation}
where $n_{fp}$ is the number of positive samples classified incorrectly (false positive). We finally compute the \textit{Geometric Mean}, which is the primary metric considered to evaluate the performance of a configuration
\begin{equation}\label{eq:metrics_gmean}
    gm = \sqrt{tpr \times tnr}.
\end{equation}

\fakepar{Earliness evaluation:} we evaluate earliness by defining four additional metrics. We first compute the \textit{Average Delay}, defined as
\begin{equation}\label{eq:average_delay}
del = \frac{1}{N_\mathcal{E}} \sum_{e \in \mathcal{E}} (t_e - \tau_{1e}),
\end{equation}
where $\mathcal{E}$ denotes the set of events in the testing split (positive instances), $N_\mathcal{E}$ is the number of events in this earliness dataset, $t_e$ is the time at which event $e$ is triggered by the model, and $\tau_{1e}$ is the true beginning of event $e$.
False positives are not included in the delay calculation: when no event occurs, the model is expected not to trigger. We therefore define the \textit{False Trigger Rate} to measure the amount of event triggers produced by the model outside the event range:
\begin{equation}
ftr = \frac{n_{ft}}{N_{\mathcal{E}}}.
\end{equation}
The complement of the \textit{False Trigger Rate} is the \textit{True Trigger Rate}, defined as
\begin{equation}
ttr = 1 - ftr.
\end{equation}
Ultimately, we define the \textit{Average Earliness} relative to the Critical Clearing Time (CCT) as
\begin{equation}\label{eq:earliness}
earl = \frac{cct - del}{cct}.
\end{equation}
If the \textit{Average Delay} exceeds the CCT, the \textit{Average Earliness} is set to $0$.

\fakepar{Baselines and comparisons:} we compare the results obtained with our new graph-embedded regularizers (7-9) with those obtained using regularizers 0-6, which have already demonstrated effectiveness with MS-SVDD. We also compare the MS-SVDD results with two external baselines: OCSVM and CCS-SVDD. Since OCSVM is not designed for multimodal data, we concatenate all modalities into a single representation to make them suitable for training. Following \cite{wang2024consistency}, we implement our own version of CCS-SVDD and fine-tune it using the same hyperparameter grid as in the original article.
Using different decision strategies, we additionally give some insights about the role of each modality in the event detection process.

\subsection{Experiment results and discussion}
All experiments are conducted on an Intel(R) Core(TM) Ultra 7 255H CPU with 32 GB of RAM.
As we work with a limited synthetic dataset, we do not consider real-world constraints such as communication latency or on-device execution time. Training and testing times are reported solely for complexity comparison purposes and are not directly included in the earliness evaluation, as our implementations could be further optimized using GPUs. In Table \ref{tab:results-noise-0} and Table \ref{tab:results-noise-10}, we report the best evaluated configurations and results. We evaluate every possible configuration across four variants of our dataset: 0\% noise linear, 0\% noise non-linear (with NPT), 10\% noise linear, and 10\% noise non-linear (with NPT). The tables present only the best configurations for each MS-SVDD regularizer and each baseline. Detailed tables of all the experiments, including the best hyperparameters for each configurations, are provided in the Supplementary Material.

\begin{table}[ht]
\centering  
\scriptsize  
\setlength{\tabcolsep}{1.5pt}  
\renewcommand{\arraystretch}{1.4}
\begin{tabular}{ccccccccccccc}
\hline
\textbf{Method}   & \multicolumn{1}{c|}{\textbf{Strat}} & \textbf{Acc} & \textbf{TPR} & \textbf{TNR} & \textbf{Pre} & \multicolumn{1}{c|}{\textbf{GM}}   & \textbf{\begin{tabular}[c]{@{}c@{}}train\\ time\end{tabular}} & \multicolumn{1}{c|}{\textbf{\begin{tabular}[c]{@{}c@{}}test\\ time\end{tabular}}} & \textbf{del} & \textbf{ftr}  & \textbf{ttr}  & \textbf{earl} \\ \hline
\multicolumn{13}{c}{\textbf{0\% Noise - Linear}}                                                                                                                                                                                                                                                                                                                            \\ \hline
OCSVM             & \multicolumn{1}{c|}{-}              & 0.46         & 0.80         & 0.12         & 0.47         & \multicolumn{1}{c|}{0.30}          & 0.002                                                         & \multicolumn{1}{c|}{0.001}                                                        & -            & 1.00          & 0.00          & 0.00          \\
CCSSVDD           & \multicolumn{1}{c|}{$\text{UNI}_1$}         & 0.49         & 0.9          & 0.09         & 0.5          & \multicolumn{1}{c|}{0.29}          & 47.158                                                        & \multicolumn{1}{c|}{0.011}                                                        & -            & 1.00          & 0.00          & 0.00          \\
MS-SVDD-$\omega_0$ & \multicolumn{1}{c|}{AND}            & 0.4          & 0.71         & 0.1          & 0.44         & \multicolumn{1}{c|}{0.26}          & 0.177                                                         & \multicolumn{1}{c|}{0.004}                                                        & -            & 1.00          & 0.00          & 0.00          \\
MS-SVDD-$\omega_1$ & \multicolumn{1}{c|}{$\text{UNI}_1$}         & 0.48         & 0.84         & 0.12         & 0.49         & \multicolumn{1}{c|}{\textbf{0.32}} & 0.336                                                         & \multicolumn{1}{c|}{0.011}                                                        & -            & 1.00          & 0.00          & 0.00          \\
MS-SVDD-$\omega_2$ & \multicolumn{1}{c|}{$\text{UNI}_1$}         & 0.47         & 0.84         & 0.1          & 0.48         & \multicolumn{1}{c|}{0.29}          & 0.403                                                         & \multicolumn{1}{c|}{0.010}                                                        & -            & 1.00          & 0.00          & 0.00          \\
MS-SVDD-$\omega_3$ & \multicolumn{1}{c|}{$\text{UNI}_1$}         & 0.48         & 0.88         & 0.07         & 0.49         & \multicolumn{1}{c|}{0.25}          & 0.491                                                         & \multicolumn{1}{c|}{0.012}                                                        & -            & 1.00          & 0.00          & 0.00          \\
MS-SVDD-$\omega_4$ & \multicolumn{1}{c|}{$\text{UNI}_1$}         & 0.46         & 0.85         & 0.07         & 0.48         & \multicolumn{1}{c|}{0.25}          & 0.328                                                         & \multicolumn{1}{c|}{0.008}                                                        & -            & 1.00          & 0.00          & 0.00          \\
MS-SVDD-$\omega_5$ & \multicolumn{1}{c|}{$\text{UNI}_1$}         & 0.45         & 0.77         & 0.13         & 0.47         & \multicolumn{1}{c|}{0.31}          & 0.445                                                         & \multicolumn{1}{c|}{0.010}                                                        & -            & 1.00          & 0.00          & 0.00          \\
MS-SVDD-$\omega_6$ & \multicolumn{1}{c|}{AND}            & 0.4          & 0.71         & 0.1          & 0.44         & \multicolumn{1}{c|}{0.26}          & 1.075                                                         & \multicolumn{1}{c|}{0.010}                                                        & -            & 1.00          & 0.00          & 0.00          \\
\textbf{MS-SVDD-$\omega_7$} & \multicolumn{1}{c|}{$\text{UNI}_1$}         & 0.47         & 0.84         & 0.11         & 0.48         & \multicolumn{1}{c|}{0.30}          & 1.049                                                         & \multicolumn{1}{c|}{0.022}                                                        & -            & 1.00          & 0.00          & 0.00          \\
\textbf{MS-SVDD-$\omega_8$} & \multicolumn{1}{c|}{$\text{UNI}_1$}         & 0.47         & 0.85         & 0.09         & 0.48         & \multicolumn{1}{c|}{0.28}          & 10.258                                                        & \multicolumn{1}{c|}{0.005}                                                        & -            & 1.00          & 0.00          & 0.00          \\
\textbf{MS-SVDD-$\omega_9$} & \multicolumn{1}{c|}{$\text{UNI}_1$}         & 0.48         & 0.84         & 0.12         & 0.49         & \multicolumn{1}{c|}{\textbf{0.32}}          & 8.237                                                         & \multicolumn{1}{c|}{0.024}                                                        & -            & 1.00          & 0.00          & 0.00          \\ \hline
\multicolumn{13}{c}{\textbf{0\% Noise - Non-linear}}                                                                                                                                                                                                                                                                                                                        \\ \hline
OCSVM             & \multicolumn{1}{c|}{-}              & 0.72         & 0.99         & 0.44         & 0.64         & \multicolumn{1}{c|}{0.66}          & 0.915                                                         & \multicolumn{1}{c|}{0.346}                                                        & 0.007        & 0.40          & 0.60          & 0.93          \\
CCSSVDD           & \multicolumn{1}{c|}{AND}            & 0.67         & 0.85         & 0.49         & 0.63         & \multicolumn{1}{c|}{0.65}          & 55.58                                                         & \multicolumn{1}{c|}{0.652}                                                        & 0.034        & 0.38          & 0.62          & 0.66          \\
MS-SVDD-$\omega_0$ & \multicolumn{1}{c|}{$\text{UNI}_1$}         & 0.76         & 0.93         & 0.58         & 0.69         & \multicolumn{1}{c|}{0.74}          & 2.303                                                         & \multicolumn{1}{c|}{0.476}                                                        & \textbf{0.003}        & 0.30          & 0.70          & \textbf{0.97}          \\
MS-SVDD-$\omega_1$ & \multicolumn{1}{c|}{$\text{UNI}_1$}         & 0.76         & 0.93         & 0.58         & 0.69         & \multicolumn{1}{c|}{0.74}          & 2.868                                                         & \multicolumn{1}{c|}{0.385}                                                        & 0.004        & 0.26          & 0.74          & 0.96          \\
MS-SVDD-$\omega_2$ & \multicolumn{1}{c|}{$\text{UNI}_1$}         & 0.8          & 0.95         & 0.65         & 0.73         & \multicolumn{1}{c|}{0.79}          & 5.856                                                         & \multicolumn{1}{c|}{0.424}                                                        & 0.014        & 0.12          & 0.88          & 0.86          \\
MS-SVDD-$\omega_3$ & \multicolumn{1}{c|}{$\text{UNI}_1$}         & 0.76         & 0.93         & 0.59         & 0.69         & \multicolumn{1}{c|}{0.74}          & 8.064                                                         & \multicolumn{1}{c|}{0.43}                                                         & \textbf{0.003}        & 0.26          & 0.74          & \textbf{0.97} \\
MS-SVDD-$\omega_4$ & \multicolumn{1}{c|}{AND}            & 0.80         & 0.84         & 0.76         & 0.78         & \multicolumn{1}{c|}{\textbf{0.80}} & 3.374                                                         & \multicolumn{1}{c|}{0.279}                                                        & 0.020        & \textbf{0.08}          & \textbf{0.92}          & 0.80          \\
MS-SVDD-$\omega_5$ & \multicolumn{1}{c|}{AND}            & 0.76         & 0.85         & 0.66         & 0.72         & \multicolumn{1}{c|}{0.75}          & 3.399                                                         & \multicolumn{1}{c|}{0.526}                                                        & 0.017        & 0.12          & 0.88          & 0.83          \\
MS-SVDD-$\omega_6$ & \multicolumn{1}{c|}{$\text{UNI}_1$}         & 0.77         & 0.93         & 0.61         & 0.71         & \multicolumn{1}{c|}{0.75}          & 5.827                                                         & \multicolumn{1}{c|}{0.387}                                                        & 0.015        & 0.16          & 0.84          & 0.85          \\
\textbf{MS-SVDD-$\omega_7$} & \multicolumn{1}{c|}{AND}            & 0.77         & 0.85         & 0.70         & 0.74         & \multicolumn{1}{c|}{0.77}          & 3.666                                                         & \multicolumn{1}{c|}{0.395}                                                        & 0.023        & \textbf{0.08}          & \textbf{0.92}          & 0.77          \\
\textbf{MS-SVDD-$\omega_8$} & \multicolumn{1}{c|}{$\text{UNI}_1$}         & 0.80         & 0.95         & 0.65         & 0.73         & \multicolumn{1}{c|}{0.78}          & 13.57                                                         & \multicolumn{1}{c|}{1.313}                                                        & 0.015        & 0.14          & 0.86          & 0.85          \\
\textbf{MS-SVDD-$\omega_9$} & \multicolumn{1}{c|}{AND}            & 0.79         & 0.84         & 0.74         & 0.77         & \multicolumn{1}{c|}{0.79}          & 11.806                                                        & \multicolumn{1}{c|}{0.876}                                                        & 0.064        & 0.10 & 0.90 & 0.36          \\ \hline
\end{tabular}
\caption{Best configurations and results for every variant on the
multimodal dataset with 0\% noise. Durations and delays are expressed in seconds.}
\label{tab:results-noise-0}
\end{table}

With linear data description, both at 0\% and 10\% noise, multimodal models do not provide any significant advantage over OCSVM. The highest \textit{Geometric Mean} score reaches $0.32$ for 0\% noise and $0.34$ for 10\% noise. While the \textit{True Positive Rate} is high, the overall \textit{Geometric Mean} is weighed down by the \textit{True Negative Rate}, which does not exceed $0.13$ for any configuration. This indicates that the configurations are overly sensitive to variations in the linear case. Consequently, the models consistently trigger outside the event range during the earliness evaluation, reducing the \textit{False Trigger Rate} to 0. This underscores the need to transform the data with NPT before addressing the OCC problem.

When applying the Non-linear Projection Trick (NPT), MS-SVDD results improve dramatically, reaching a \textit{Geometric Mean} of $0.80$ for 0\% noise and $0.78$ for 10\% noise. These higher scores are explained by the increase in specificity, with the MS-SVDD \textit{True Negative Rate} now ranging between $0.58$ and $0.76$ for 0\% noise, and between $0.45$ and $0.68$ for 10\% noise. For both variants, MS-SVDD also demonstrates better reliability than OCSVM and CSS-SVDD, with an improvement of at least $0.09$ between the worst MS-SVDD configuration and the CSS-SVDD implementation. We notice that under noise, OCSVM is unable to solve the OCC problem.

\begin{table}[ht]
\centering  
\scriptsize  
\setlength{\tabcolsep}{1.5pt}  
\renewcommand{\arraystretch}{1.4}
\begin{tabular}{ccccccccccccc}
\hline
\textbf{Method}   & \multicolumn{1}{c|}{\textbf{Strat}} & \textbf{Acc} & \textbf{TPR} & \textbf{TNR} & \textbf{Pre} & \multicolumn{1}{c|}{\textbf{GM}}   & \textbf{\begin{tabular}[c]{@{}c@{}}train\\ time\end{tabular}} & \multicolumn{1}{c|}{\textbf{\begin{tabular}[c]{@{}c@{}}test\\ time\end{tabular}}} & \textbf{del} & \textbf{ftr} & \textbf{ttr} & \textbf{earl} \\ \hline
\multicolumn{13}{c}{\textbf{10\% Noise - Linear}}                                                                                                                                                                                                                                                                                                                         \\ \hline
OCSVM             & \multicolumn{1}{c|}{}               & 0.48         & 0.83         & 0.14         & 0.49         & \multicolumn{1}{c|}{\textbf{0.34}} & 0.003                                                         & \multicolumn{1}{c|}{0.002}                                                        & -            & 1.00         & 0.00         & 0.00          \\
CCSSVDD           & \multicolumn{1}{c|}{$\text{UNI}_1$}         & 0.52         & 0.92         & 0.11         & 0.51         & \multicolumn{1}{c|}{0.32}          & 38.951                                                        & \multicolumn{1}{c|}{0.011}                                                        & -            & 1.00         & 0.00         & 0.00          \\
MS-SVDD-$\omega_0$ & \multicolumn{1}{c|}{AND}            & 0.41         & 0.71         & 0.11         & 0.44         & \multicolumn{1}{c|}{0.28}          & 0.124                                                         & \multicolumn{1}{c|}{0.003}                                                        & -            & 1.00         & 0.00         & 0.00          \\
MS-SVDD-$\omega_1$ & \multicolumn{1}{c|}{$\text{UNI}_1$}         & 0.48         & 0.83         & 0.12         & 0.49         & \multicolumn{1}{c|}{0.32}          & 0.482                                                         & \multicolumn{1}{c|}{0.010}                                                        & -            & 1.00         & 0.00         & 0.00          \\
MS-SVDD-$\omega_2$ & \multicolumn{1}{c|}{AND}            & 0.41         & 0.72         & 0.11         & 0.45         & \multicolumn{1}{c|}{0.28}          & 0.456                                                         & \multicolumn{1}{c|}{0.011}                                                        & -            & 1.00         & 0.00         & 0.00          \\
MS-SVDD-$\omega_3$ & \multicolumn{1}{c|}{AND}            & 0.41         & 0.71         & 0.11         & 0.44         & \multicolumn{1}{c|}{0.28}          & 0.249                                                         & \multicolumn{1}{c|}{0.007}                                                        & -            & 1.00         & 0.00         & 0.00          \\
MS-SVDD-$\omega_4$ & \multicolumn{1}{c|}{$\text{UNI}_1$}         & 0.48         & 0.83         & 0.13         & 0.49         & \multicolumn{1}{c|}{0.33}          & 0.425                                                         & \multicolumn{1}{c|}{0.010}                                                        & -            & 1.00         & 0.00         & 0.00          \\
MS-SVDD-$\omega_5$ & \multicolumn{1}{c|}{AND}            & 0.41         & 0.72         & 0.11         & 0.45         & \multicolumn{1}{c|}{0.28}          & 0.406                                                         & \multicolumn{1}{c|}{0.010}                                                        & -            & 1.00         & 0.00         & 0.00          \\
MS-SVDD-$\omega_6$ & \multicolumn{1}{c|}{AND}            & 0.41         & 0.71         & 0.11         & 0.44         & \multicolumn{1}{c|}{0.28}          & 1.071                                                         & \multicolumn{1}{c|}{0.007}                                                        & -            & 1.00         & 0.00         & 0.00          \\
\textbf{MS-SVDD-$\omega_7$} & \multicolumn{1}{c|}{$\text{UNI}_1$}         & 0.48         & 0.85         & 0.12         & 0.49         & \multicolumn{1}{c|}{0.32}          & 0.849                                                         & \multicolumn{1}{c|}{0.007}                                                        & -            & 1.00         & 0.00         & 0.00          \\
\textbf{MS-SVDD-$\omega_8$} & \multicolumn{1}{c|}{$\text{UNI}_1$}         & 0.48         & 0.85         & 0.10         & 0.49         & \multicolumn{1}{c|}{0.29}          & 9.580                                                         & \multicolumn{1}{c|}{0.012}                                                        & -            & 1.00         & 0.00         & 0.00          \\
\textbf{MS-SVDD-$\omega_9$} & \multicolumn{1}{c|}{$\text{UNI}_1$}         & 0.49         & 0.85         & 0.12         & 0.49         & \multicolumn{1}{c|}{0.32}          & 8.035                                                         & \multicolumn{1}{c|}{0.027}                                                        & -            & 1.00         & 0.00         & 0.00          \\ \hline
\multicolumn{13}{c}{\textbf{10\% Noise - Non-linear}}                                                                                                                                                                                                                                                                                                                     \\ \hline
OCSVM             & \multicolumn{1}{c|}{}               & 0.50         & 1.00         & 0.00         & 0.50         & \multicolumn{1}{c|}{0.00}          & 0.991                                                         & \multicolumn{1}{c|}{0.152}                                                        & 0.092        & 0.94         & 0.06         & 0.08          \\
CCSSVDD           & \multicolumn{1}{c|}{AND}            & 0.53         & 0.84         & 0.23         & 0.52         & \multicolumn{1}{c|}{0.44}          & 60.956                                                        & \multicolumn{1}{c|}{1.423}                                                        & 0.028        & 0.90         & 0.10         & 0.72          \\
MS-SVDD-$\omega_0$ & \multicolumn{1}{c|}{AND}            & 0.67         & 0.88         & 0.46         & 0.62         & \multicolumn{1}{c|}{0.64}          & 3.477                                                         & \multicolumn{1}{c|}{0.597}                                                        & 0.013        & 0.72         & 0.28         & 0.87          \\
MS-SVDD-$\omega_1$ & \multicolumn{1}{c|}{$\text{UNI}_1$}         & 0.71         & 0.95         & 0.47         & 0.64         & \multicolumn{1}{c|}{0.67}          & 2.772                                                         & \multicolumn{1}{c|}{0.765}                                                        & \textbf{0.009}        & 0.62         & 0.38         & \textbf{0.91} \\
MS-SVDD-$\omega_2$ & \multicolumn{1}{c|}{AND}            & 0.67         & 0.90         & 0.45         & 0.62         & \multicolumn{1}{c|}{0.63}          & 4.528                                                         & \multicolumn{1}{c|}{0.773}                                                        & 0.015        & 0.76         & 0.24         & 0.85          \\
MS-SVDD-$\omega_3$ & \multicolumn{1}{c|}{$\text{UNI}_1$}         & 0.73         & 0.93         & 0.52         & 0.66         & \multicolumn{1}{c|}{0.70}          & 5.040                                                         & \multicolumn{1}{c|}{0.509}                                                        & 0.011        & 0.64         & 0.36         & 0.89          \\
MS-SVDD-$\omega_4$ & \multicolumn{1}{c|}{$\text{UNI}_1$}         & 0.70         & 0.96         & 0.44         & 0.63         & \multicolumn{1}{c|}{0.65}          & 3.889                                                         & \multicolumn{1}{c|}{0.509}                                                        & 0.011        & 0.66         & 0.34         & 0.89          \\
MS-SVDD-$\omega_5$ & \multicolumn{1}{c|}{AND}            & 0.72         & 0.89         & 0.54         & 0.66         & \multicolumn{1}{c|}{0.70}          & 3.125                                                         & \multicolumn{1}{c|}{0.378}                                                        & 0.016        & 0.62         & 0.38         & 0.84          \\
MS-SVDD-$\omega_6$ & \multicolumn{1}{c|}{$\text{UNI}_1$}         & 0.70         & 0.94         & 0.45         & 0.63         & \multicolumn{1}{c|}{0.65}          & 5.989                                                         & \multicolumn{1}{c|}{0.522}                                                        & 0.011        & 0.64         & 0.36         & 0.89          \\
\textbf{MS-SVDD-$\omega_7$} & \multicolumn{1}{c|}{$\text{UNI}_1$}         & 0.78         & 0.94         & 0.63         & 0.72         & \multicolumn{1}{c|}{0.77} & 3.867                                                         & \multicolumn{1}{c|}{0.755}                                                        & 0.035        & 0.42         & 0.58         & 0.65          \\
\textbf{MS-SVDD-$\omega_8$} & \multicolumn{1}{c|}{$\text{UNI}_1$}         & 0.77         & 0.93         & 0.62         & 0.71         & \multicolumn{1}{c|}{0.76} & 14.704                                                        & \multicolumn{1}{c|}{0.849}                                                        & 0.035        & \textbf{0.32}         & \textbf{0.68}         & 0.65          \\
\textbf{MS-SVDD-$\omega_9$} & \multicolumn{1}{c|}{AND}            & 0.79         & 0.89         & 0.68         & 0.74         & \multicolumn{1}{c|}{\textbf{0.78}} & 11.878                                                        & \multicolumn{1}{c|}{1.488}                                                        & 0.041        & 0.38         & 0.62         & 0.59          \\ \hline
\end{tabular}
\caption{Best configurations and results for every variant on the
multimodal dataset with 10\% noise. Durations and delays are expressed in seconds.}
\label{tab:results-noise-10}
\end{table}

Graph-embedded regularizers do not show any improvement at $0\%$ noise, but under noise they yield a reliability increase of between $0.06$ and $0.15$ compared to classical regularizers. This demonstrates that graph-embedded regularizers effectively leverage graph information in the data to ensure robustness in unstable environments. These regularizers are thus well designed to handle real-world variations, at the cost of a slightly longer inference time.

MS-SVDD configurations also outperform the baselines in terms of earliness in both noisy and non-noisy environments, with an \textit{Average Earliness} above $0.90$ for the best configurations, and an \textit{Average Delay} remaining below the CCT. In particular, graph-embedded regularizers generally exhibit the lowest \textit{False Trigger Rate}, with an improvement of at least $0.20$ under noise compared to other configurations (while the baselines show a high \textit{False Positive Rate} above $0.90$). This supports the conclusion that graph-embedded regularizers also improve robustness to noise with respect to earliness.

We observe that the best configurations all employ either an AND or an $\text{UNI}_1$ decision strategy. This suggests that modality 1 (reactive power) may contain more informative features for anomaly detection in this context, providing additional insight into the relative importance of each modality. The absence of OR strategies also confirms that relying on the positive classification of one random modality is insufficient, and that information from all modalities must be leveraged.

As stated in Section \ref{sec:complexity}, graph-embedded regularizers introduce additional computational complexity compared to classical regularizers, which is reflected in the testing times reported in our results. This additional cost must be balanced against the increase in robustness provided by these regularizers in order to achieve the best overall trade-off.

\section{Conclusions}

In summary, our study addresses the problem of early anomaly detection in smart power grids by implementing an extended version of the MS-SVDD algorithm. The key contributions of this work include the integration of new graph-embedded regularizers, which enable the model to leverage relationships between data points and improve the representation of multimodal information. They also include the generalization of MS-SVDD to incorporate more than two modalities, demonstrating the versatility of this model by applying it to a new type of problem.

The application of this framework to the PSML dataset, which includes voltage, active power, and reactive power measurements, demonstrates the model's ability to identify anomalies. The evaluation confirms that it can provide accurate and timely predictions, which is crucial for maintaining grid stability. Through our experiments, we show that leveraging graph information can help increase model robustness in OCC problems, especially under noise.

The ability to provide early and reliable anomaly detection supports preventive decision-making and improves grid resilience, which is essential in energy systems with a high share of renewable sources. Beyond power grids, the generalization of MS-SVDD to multiple modalities opens perspectives for other domains requiring the monitoring of heterogeneous data, such as healthcare, cybersecurity, or industrial process monitoring. Overall, our work highlights the applicability of MS-SVDD to a broader range of one-class classification problems in high-dimensional and imbalanced settings.

\bibliographystyle{elsarticle-num}
\bibliography{bibliography}

@article{Damodaram,
  title={Recent Decade Global Trends in Renewable Energy and Investments-A Review.},
  author={Damodaram, AK and Tulasi, Ch Lakshmi and Reddy, L Venkateswara},
  journal={International Journal of COMADEM},
  volume={25},
  number={3},
  year={2022}
}

@article{RUSSO2023136997,
title = {Future perspectives for wind and solar electricity production under high-resolution climate change scenarios},
journal = {Journal of Cleaner Production},
volume = {404},
pages = {136997},
year = {2023},
issn = {0959-6526},
author = {M.A. Russo and D. Carvalho and N. Martins and A. Monteiro}
}

@incollection{Berraies2021,
  title={Machine Learning to Facilitate the Integration of Renewable Energies into the Grid},
  author={Berraies, Ahlem Aissa and Tzanetos, Alexandros and Blondin, Maude},
  booktitle={Handbook of Smart Energy Systems},
  pages={689--711},
  year={2023},
  publisher={Springer}
}

@inproceedings{Penmetsa2019,
  title={Climate change effects on solar, wind and hydro power generation},
  author={Penmetsa, Vikramaditya and Holbert, Keith E},
  booktitle={2019 North American Power Symposium (NAPS)},
  pages={1--6},
  year={2019},
  organization={IEEE}
}

@inproceedings{Aula2013,
  title={Power fluctuation reduction methodology for the grid-connected renewable power systems},
  author={Aula, Fadhil T and Lee, Samuel C},
  booktitle={Active and Passive Smart Structures and Integrated Systems 2013},
  volume={8688},
  pages={565--569},
  year={2013},
  organization={SPIE}
}

@inproceedings{Liang2023101,
  title={Tutorial on multimodal machine learning: principles, challenges, and open questions},
  author={Liang, Paul Pu and Morency, Louis-Philippe},
  booktitle={Companion Publication of the 25th International Conference on Multimodal Interaction},
  pages={101--104},
  year={2023}
}

@article{zheng2021psml,
  title={A multi-scale time-series dataset with benchmark for machine learning in decarbonized energy grids},
  author={Zheng, Xiangtian and Xu, Nan and Trinh, Loc and Wu, Dongqi and Huang, Tong and Sivaranjani, S and Liu, Yan and Xie, Le},
  journal={Scientific Data},
  volume={9},
  number={1},
  pages={359},
  year={2022},
  publisher={Nature Publishing Group UK London}
}

@ARTICLE{6099519,
  author={Fang, Xi and Misra, Satyajayant and Xue, Guoliang and Yang, Dejun},
  journal={IEEE Communications Surveys \& Tutorials}, 
  title={Smart Grid — The New and Improved Power Grid: A Survey}, 
  year={2012},
  volume={14},
  number={4},
  pages={944-980},
}

@ARTICLE{6808416,
  author={Xie, Le and Chen, Yang and Kumar, P. R.},
  journal={IEEE Transactions on Power Systems}, 
  title={Dimensionality Reduction of Synchrophasor Data for Early Event Detection: Linearized Analysis}, 
  year={2014},
  volume={29},
  number={6},
  pages={2784-2794},
}

@article{SABER20201113,
title = {A threshold free PMU-based fault location scheme for multi-end lines},
journal = {Ain Shams Engineering Journal},
volume = {11},
number = {4},
pages = {1113-1121},
year = {2020},
issn = {2090-4479},
author = {Ahmed Saber and Ahmed Emam and Hany Elghazaly},
}

@inproceedings{Wu2019,
  title={Structural Analysis Based Method for Estimating Critical Clearing Time without Time Domain Simulation},
  author={Wu, D and Sharma, D and Ji, G and Perumalla, V and Luo, X and Jiang, JN},
  booktitle={2019 IEEE Power \& Energy Society General Meeting (PESGM)},
  pages={1--5},
  year={2019},
  organization={IEEE}
}

@inproceedings{Banjar-Nahor20181183,
  title={Critical clearing time transformation upon renewables integration through static converters, a case in microgrids},
  author={Banjar-Nahor, Kevin M and Garbuio, Lauric and Debusschere, Vincent and Hadjsaid, Nouredine and Sinisuka, Ngapuli and others},
  booktitle={2018 IEEE International Conference on Industrial Technology (ICIT)},
  pages={1183--1188},
  year={2018},
  organization={IEEE}
}

@ARTICLE{9207873,
  author={Gupta, Ashish and Gupta, Hari Prabhat and Biswas, Bhaskar and Dutta, Tanima},
  journal={IEEE Transactions on Artificial Intelligence}, 
  title={Approaches and Applications of Early Classification of Time Series: A Review}, 
  year={2020},
  volume={1},
  number={1},
  pages={47-61},
 }

@article{YU201937,
title = {Clustering-based proxy measure for optimizing one-class classifiers},
journal = {Pattern Recognition Letters},
volume = {117},
pages = {37-44},
year = {2019},
issn = {0167-8655},
author = {Jaehong Yu and Seokho Kang}}

@article{POLLITT201232,
title = {Lessons from the history of independent system operators in the energy sector},
journal = {Energy Policy},
volume = {47},
pages = {32-48},
year = {2012},
issn = {0301-4215},
author = {Michael G. Pollitt}}

@incollection{life_cycle_cost_assessment,
  title={Life cycle cost and life cycle assessment: an approximation to understand the real impacts of the Electricity Supply Industry},
  author={Niembro-Garc{\'\i}a, Joaquina and Alfaro-Mart{\'\i}nez, Patricia and Marmolejo-Saucedo, Jose Antonio},
  booktitle={Advances of Artificial Intelligence in a Green Energy Environment},
  pages={83--110},
  year={2022},
  publisher={Elsevier}
}

@article{SOHRAB2021107648,
title = {Multimodal subspace support vector data description},
journal = {Pattern Recognition},
volume = {110},
pages = {107648},
year = {2021},
issn = {0031-3203},
author = {Fahad Sohrab and Jenni Raitoharju and Alexandros Iosifidis and Moncef Gabbouj}
}

@INPROCEEDINGS{8545819,
  author={Sohrab, Fahad and Raitoharju, Jenni and Gabbouj, Moncef and Iosifidis, Alexandros},
  booktitle={2018 24th International Conference on Pattern Recognition (ICPR)}, 
  title={Subspace Support Vector Data Description}, 
  year={2018},
  volume={},
  number={},
  pages={722-727} }

@article{Tax2004SupportVD,
  title={Support Vector Data Description},
  author={David M. J. Tax and Robert P. W. Duin},
  journal={Machine Learning},
  year={2004},
  volume={54},
  pages={45-66}

}

@INPROCEEDINGS{10081907,
  author={Degerli, Aysen and Sohrab, Fahad and Kiranyaz, Serkan and Gabbouj, Moncef},
  booktitle={2022 Computing in Cardiology (CinC)}, 
  title={Early Myocardial Infarction Detection with One-Class Classification over Multi-view Echocardiography}, 
  year={2022},
  volume={498},
  number={},
  pages={1-4}
 }

@InProceedings{trustworthiness2024,
author="Khan, Tanveer
and Sohrab, Fahad
and Michalas, Antonis
and Gabbouj, Moncef",
editor="Barolli, Leonard",
title="Trustworthiness of X-Users: A One-Class Classification Approach",
booktitle="Advanced Information Networking and Applications",
year="2024",
publisher="Springer Nature Switzerland",
address="Cham",
pages="331--343",
}

@INPROCEEDINGS{10372038,
  author={Zaffar, Zaffar and Sohrab, Fahad and Kanniainen, Juho and Gabbouj, Moncef},
  booktitle={2023 IEEE Symposium Series on Computational Intelligence (SSCI)}, 
  title={Credit Card Fraud Detection with Subspace Learning-based One-Class Classification}, 
  year={2023},
  pages={407-412}
}

@article{KIM2019353,
title = {Bus voltage control and optimization strategies for power flow analyses using Petri net approach},
journal = {International Journal of Electrical Power \& Energy Systems},
volume = {112},
pages = {353-361},
year = {2019},
issn = {0142-0615},
author = {Insu Kim and Shuo Xu}}

@ARTICLE{6584012,
  author={Kwak, Nojun},
  journal={IEEE Transactions on Neural Networks and Learning Systems}, 
  title={Nonlinear Projection Trick in Kernel Methods: An Alternative to the Kernel Trick}, 
  year={2013},
  volume={24},
  number={12},
  pages={2113-2119}}

@inproceedings{Han20244591,
  title={ERL-MR: Harnessing the Power of Euler Feature Representations for Balanced Multi-modal Learning},
  author={Han, Weixiang and Cai, Chengjun and Guo, Yu and Peng, Jialiang},
  booktitle={Proceedings of the 32nd ACM International Conference on Multimedia},
  pages={4591--4600},
  year={2024}
}

@article{zheng2016smoothly,
  title={Smoothly approximated support vector domain description},
  author={Zheng, Songfeng},
  journal={Pattern Recognition},
  volume={49},
  pages={55--64},
  year={2016},
  publisher={Elsevier}
}

@inproceedings{zahid2024refining,
  title={Refining Myocardial Infarction Detection: A Novel Multi-Modal Composite Kernel Strategy in One-Class Classification},
  author={Zahid, Muhammad Uzair and Degerli, Aysen and Sohrab, Fahad and Kiranyaz, Serkan and Hamid, Tahir and Mazhar, Rashid and Gabbouj, Moncef},
  booktitle={2024 IEEE International Conference on Image Processing (ICIP)},
  pages={3010--3016},
  year={2024},
  organization={IEEE}
}

@article{sohrab2020ellipsoidal,
  title={Ellipsoidal subspace support vector data description},
  author={Sohrab, Fahad and Raitoharju, Jenni and Iosifidis, Alexandros and Gabbouj, Moncef},
  journal={IEEE Access},
  volume={8},
  pages={122013--122025},
  year={2020},
  publisher={IEEE}
}

@article{al2024malware,
  title={Malware Detection with Subspace Learning-based One-Class Classification},
  author={Al-Khshali, Hasan H and Ilyas, Muhammad and Sohrab, Fahad and Gabbouj, Moncef},
  journal={IEEE Access},
  year={2024},
  publisher={IEEE}
}

@inproceedings{kilickaya2023hyperspectral,
  title={Hyperspectral image analysis with subspace learning-based one-class classification},
  author={Kilickaya, Sertac and Ahishali, Mete and Sohrab, Fahad and Ince, Turker and Gabbouj, Moncef},
  booktitle={2023 Photonics \& Electromagnetics Research Symposium (PIERS)},
  pages={953--959},
  year={2023},
  organization={IEEE}
}

@article{laakom2023convolutional,
  title={Convolutional autoencoder-based multimodal one-class classification},
  author={Laakom, Firas and Sohrab, Fahad and Raitoharju, Jenni and Iosifidis, Alexandros and Gabbouj, Moncef},
  journal={arXiv preprint arXiv:2309.14090},
  year={2023}
}

@article{yang2019one,
  title={One-class classification using generative adversarial networks},
  author={Yang, Yang and Hou, Chunping and Lang, Yue and Yue, Guanghui and He, Yuan},
  journal={IEEE Access},
  volume={7},
  pages={37970--37979},
  year={2019},
  publisher={IEEE}
}

@article{tsai2021feature,
  title={Feature selection and ensemble learning techniques in one-class classifiers: an empirical study of two-class imbalanced datasets},
  author={Tsai, Chih-Fong and Lin, Wei-Chao},
  journal={IEEE Access},
  volume={9},
  pages={13717--13726},
  year={2021},
  publisher={IEEE}
}

@article{FSVDD2011,
  title={An extension to fuzzy support vector data description (FSVDD*)},
  author={Forghani, Yahya and Sadoghi Yazdi, H and Effati, Sohrab},
  journal={Pattern Analysis and Applications},
  volume={15},
  number={3},
  pages={237--247},
  year={2012},
  publisher={Springer}
}

@article{zuhaib2025identification,
  title={Identification and suppression of low-frequency oscillations using PMU measurements based power system model in smart grid},
  author={Zuhaib, Mohd and Rihan, Mohd and Gupta, Saket and Sufyan, Marwan Ahmad Abdullah},
  journal={Scientific Reports},
  volume={15},
  number={1},
  pages={3822},
  year={2025},
  publisher={Nature Publishing Group UK London}
}

@article{choobdari2024robust,
  title={Robust distribution networks reconfiguration considering the improvement of network resilience considering renewable energy resources},
  author={Choobdari, Mahsa and Samiei Moghaddam, Mahmoud and Davarzani, Reza and Azarfar, Azita and Hoseinpour, Hesamodin},
  journal={Scientific Reports},
  volume={14},
  number={1},
  pages={23041},
  year={2024},
  publisher={Nature Publishing Group UK London}
}

@article{rajak2025multiobjective,
  title={Multiobjective adaptive predictive virtual synchronous generator control strategy for grid stability and renewable integration},
  author={Rajak, Mrinal Kanti and Pudur, Rajen},
  journal={Scientific Reports},
  volume={15},
  number={1},
  pages={9241},
  year={2025},
  publisher={Nature Publishing Group UK London}
}

@article{wang2026distribution,
  title={Distribution entropy regularized multimodal subspace support vector data description for anomaly detection},
  author={Wang, Chuang and Ning, Xin and Qian, Pengjiang and Hu, Wenjun and Yao, Jian and Ng, Eddie-Yin-Kwee and Lai, Khin Wee and Wang, Shitong},
  journal={Pattern Recognition},
  volume={172},
  pages={112478},
  year={2026},
  publisher={Elsevier},
  doi={10.1016/j.patcog.2025.112478}
}

@article{wang2024consistency,
  title={Consistency and Complementarity Jointly Regularized Subspace Support Vector Data Description for Multimodal Data},
  author={Wang, Chuang and Hu, Wenjun and Wang, Juan and Qian, Pengjiang and Wang, Shitong and Ortale, Riccardo},
  journal={International Journal of Intelligent Systems},
  volume={2024},
  pages={1--16},
  year={2024},
  publisher={Wiley},
  doi={10.1155/2024/1989706}
}

\journal{pattern recognition}

\newpage
\setcounter{figure}{0}  
\setcounter{table}{0} 
\setcounter{section}{0}  
\setcounter{page}{1}  
\begin{center}\textbf{Anomaly Detection in Smart Power Grids with Graph-Regularized MS-SVDD \\ Supplementary Material}
\end{center}

\begin{center}
Thomas Debelle, Fahad Sohrab, Pekka Abrahamsson, Moncef Gabbouj
\end{center}

This document contains supplementary material for the proposed Graph-Regularized Multi-modal Subspace Support Vector Data Description (MS-SVDD). Section 1 provides a chart describing the preprocessing steps to create the reliability dataset. In Section 2, we report all the
experimental results for MS-SVDD and CCS-SVDD, as well as the best hyperparameters found.

\section{Preprocessing of the PSML dataset for reliability evaluation}\label{sec:preprocessing}
\begin{figure}[H]   
	\centering
    \includegraphics[scale=0.5]{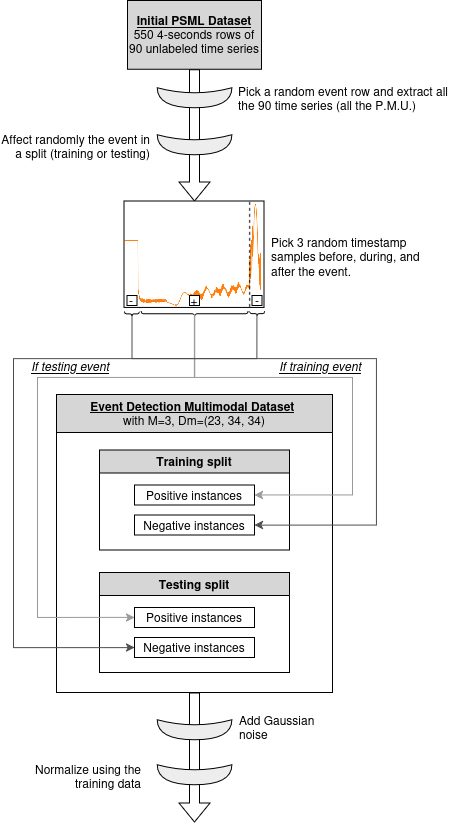}
	\caption{Preprocessing of PSML dataset for reliability evaluation.}
	\label{fig:preprocessing}
\end{figure}

\newpage
\section{Full results of the experiments}

\begin{table}[H]
\centering  
\scriptsize  
\setlength{\tabcolsep}{3pt}  
\renewcommand{\arraystretch}{1.0}
\begin{tabular}{ccc|ccccc|ccccc|cc}
\hline
\textbf{Strat} & \textbf{NPT} & \textbf{Noise} & \textbf{$\sigma$} & \textbf{$C$} & \textbf{$\beta$} & \textbf{$\eta$} & \textbf{$d$} & \textbf{Acc} & \textbf{TPR} & \textbf{TNR} & \textbf{Pre} & \textbf{GM} & \textbf{\begin{tabular}[c]{@{}c@{}}train\\ time\end{tabular}} & \textbf{\begin{tabular}[c]{@{}c@{}}test\\ time\end{tabular}} \\ \hline
AND            & 0            & 0              & 0                 & 0.02         & 10.0             & 0.1             & 10           & 0.47         & 0.85         & 0.09         & 0.48         & 0.28        & 39.572                                                        & 0.010                                                        \\
AND            & 0            & 1              & 0                 & 0.02         & 1.0              & 0.1             & 10           & 0.48         & 0.86         & 0.11         & 0.49         & 0.31        & 39.607                                                        & 0.010                                                        \\
AND            & 1            & 0              & 0.01              & 0.01         & 100.0            & 0.1             & 10           & 0.67         & 0.85         & 0.49         & 0.63         & 0.65        & 55.580                                                        & 0.652                                                        \\
AND            & 1            & 1              & 0.01              & 0.01         & 100.0            & 0.1             & 10           & 0.53         & 0.84         & 0.23         & 0.52         & 0.44        & 60.956                                                        & 1.423                                                        \\
OR             & 0            & 0              & 0                 & 0.02         & 10.0             & 0.1             & 10           & 0.50         & 1.00         & 0.01         & 0.50         & 0.08        & 38.641                                                        & 0.011                                                        \\
OR             & 0            & 1              & 0                 & 0.02         & 0.01             & 0.1             & 15           & 0.50         & 1.00         & 0.00         & 0.50         & 0.00        & 40.117                                                        & 0.014                                                        \\
OR             & 1            & 0              & 0.01              & 0.01         & 100.0            & 0.1             & 10           & 0.50         & 1.00         & 0.00         & 0.50         & 0.00        & 55.205                                                        & 0.468                                                        \\
OR             & 1            & 1              & 0.01              & 0.01         & 100.0            & 0.1             & 10           & 0.50         & 0.99         & 0.00         & 0.50         & 0.00        & 61.283                                                        & 0.644                                                        \\
$\text{UNI}_0$         & 0            & 0              & 0                 & 0.02         & 10.0             & 0.1             & 10           & 0.47         & 0.85         & 0.09         & 0.48         & 0.28        & 39.722                                                        & 0.011                                                        \\
$\text{UNI}_0$         & 0            & 1              & 0                 & 0.02         & 1.0              & 0.1             & 10           & 0.48         & 0.86         & 0.11         & 0.49         & 0.31        & 39.032                                                        & 0.011                                                        \\
$\text{UNI}_0$         & 1            & 0              & 0.01              & 0.01         & 100.0            & 0.1             & 10           & 0.67         & 0.85         & 0.49         & 0.63         & 0.65        & 55.566                                                        & 0.487                                                        \\
$\text{UNI}_0$         & 1            & 1              & 0.01              & 0.01         & 100.0            & 0.1             & 10           & 0.53         & 0.84         & 0.23         & 0.52         & 0.44        & 60.874                                                        & 0.673                                                        \\
$\text{UNI}_1$         & 0            & 0              & 0                 & 0.02         & 10.0             & 0.1             & 10           & 0.49         & 0.90         & 0.09         & 0.50         & 0.29        & 47.158                                                        & 0.011                                                        \\
$\text{UNI}_1$         & 0            & 1              & 0                 & 0.02         & 1.0              & 0.1             & 10           & 0.52         & 0.92         & 0.11         & 0.51         & 0.32        & 38.951                                                        & 0.011                                                        \\
$\text{UNI}_1$         & 1            & 0              & 0.01              & 0.01         & 100.0            & 0.1             & 10           & 0.67         & 0.98         & 0.36         & 0.60         & 0.59        & 56.016                                                        & 0.485                                                        \\
$\text{UNI}_1$         & 1            & 1              & 0.01              & 0.01         & 100.0            & 0.1             & 10           & 0.56         & 0.96         & 0.16         & 0.53         & 0.39        & 61.421                                                        & 0.718                                                        \\
$\text{UNI}_2$         & 0            & 0              & 0                 & 0.02         & 10.0             & 0.1             & 10           & 0.49         & 0.90         & 0.09         & 0.50         & 0.29        & 47.140                                                        & 0.011                                                        \\
$\text{UNI}_2$         & 0            & 1              & 0                 & 0.02         & 1.0              & 0.1             & 10           & 0.52         & 0.92         & 0.11         & 0.51         & 0.32        & 39.967                                                        & 0.011                                                        \\
$\text{UNI}_2$         & 1            & 0              & 0.01              & 0.01         & 100.0            & 0.1             & 10           & 0.67         & 0.98         & 0.36         & 0.60         & 0.59        & 61.586                                                        & 0.495                                                        \\
$\text{UNI}_2$         & 1            & 1              & 0.01              & 0.01         & 100.0            & 0.1             & 10           & 0.56         & 0.96         & 0.16         & 0.53         & 0.39        & 61.890                                                        & 0.72                                                         \\ \hline
\end{tabular}
\caption{Full results for CCS-SVDD experiments.}
\label{tab:full-results-CCSSVDD}
\end{table}

\centering  
\scriptsize
\setlength{\tabcolsep}{1.5pt}  
\renewcommand{\arraystretch}{1.0}
\begin{longtable}{cccc|cccccc|ccccc|cc}
\hline
\textbf{Reg.} & \textbf{Strat} & \textbf{NPT} & \textbf{Noise} & \textbf{$\sigma$} & \textbf{$k$} & \textbf{$C$} & \textbf{$\beta$} & \textbf{$\eta$} & \textbf{$d$} & \textbf{Acc} & \textbf{TPR} & \textbf{TNR} & \textbf{Pre} & \textbf{GM} & \textbf{\begin{tabular}[c]{@{}c@{}}train\\ time\end{tabular}} & \textbf{\begin{tabular}[c]{@{}c@{}}test\\ time\end{tabular}} \\ \hline
0             & AND            & 0            & 0              & 0                 & 0            & 0.01         & 0.001            & 0.1             & 5            & 0.40         & 0.71         & 0.10         & 0.44         & 0.26        & 0.177                                                         & 0.004                                                        \\
0             & AND            & 0            & 1              & 0                 & 0            & 0.01         & 0.01             & 0.1             & 2            & 0.41         & 0.71         & 0.11         & 0.44         & 0.28        & 0.124                                                         & 0.003                                                        \\
0             & AND            & 1            & 0              & 0.01              & 0            & 0.01         & 1.0              & 0.1             & 5            & 0.72         & 0.85         & 0.58         & 0.67         & 0.70        & 2.753                                                         & 0.509                                                        \\
0             & AND            & 1            & 1              & 0.01              & 0            & 0.01         & 10000.0          & 0.1             & 4            & 0.67         & 0.88         & 0.46         & 0.62         & 0.64        & 3.477                                                         & 0.597                                                        \\
\end{longtable}
\end{document}